\documentclass[journal]{IEEEtran}

\usepackage{generic}
\usepackage{cite}
\usepackage{amsmath,amssymb,amsfonts}
\usepackage{algorithmic}
\usepackage{textcomp}
\usepackage[table,xcdraw]{xcolor}
\usepackage{amsmath}
\usepackage{mathabx}
\usepackage{rotating}
\usepackage{graphicx}
\usepackage{booktabs}
\usepackage{tabularx}
\usepackage[lofdepth,lotdepth]{subfig}
\usepackage{lscape}

\begin{document}

\title{Online Non-Destructive Moisture Content Estimation of Filter Media During Drying Using Artificial Neural Networks}

\author{Christian~Remi~Wewer~and~Alexandros~Iosifidis\\~\textit{Department of Electrical and Computer Engineering, Aarhus~University,~Denmark\\
\{chw,ai\}@ece.au.dk}}

\maketitle

\renewcommand{\arraystretch}{1.2}

\begin{abstract}
Moisture content (MC) estimation is important in the manufacturing process of drying bulky filter media products as it is the prerequisite for drying optimization. 
In this study, a dataset collected by performing 161 drying industrial experiments is described and a methodology for MC estimation in an non-destructive and online manner during industrial drying is presented. An artificial neural network (ANN) based method is compared to state-of-the-art MC estimation methods reported in the literature. Results of model fitting and training show that a three-layer Perceptron achieves the lowest error. 
Experimental results show that ANNs combined with oven settings data, drying time and product temperature can be used to reliably estimate the MC of bulky filter media products.

\end{abstract}

\begin{IEEEkeywords}
Drying, Artificial Neural Networks, Moisture Content, Moisture Content Prediction, Moisture Content Estimation, Filter Media.
\end{IEEEkeywords}

\section{Introduction}

Drying is a widely used manufacturing process across many different fields of manufacturing. 
For the manufacturing of filter media, the drying process is the most energy intensive and time consuming process.

The drying of filter media products is a process highly dependent on both the upstream manufacturing steps and the state of the ambient air, due to the hygroscopic nature of the filter media. 
These dependencies introduce variance in the MC of the filter media before the drying process and thus also introduce variance in the required drying times to reach a desired MC threshold for each filter media product. 
Therefore, control of the drying process of filter media products is based on a time threshold ensuring that all filters reach a desired MC threshold. This process comes at the cost of energy and time which is spent over-drying filter media products.

However, with the knowledge of MC of filter media products, it is possible to reduce both drying time and energy expenditure. This is because it is possible to stop the individual drying processes based on the condition of the filter media, instead of a predetermined time threshold. 
Direct measurement of the filter media MC during the drying process is infeasible, introducing the need of alternative methods.

One approach is to model the drying of wet materials. This is a complex, highly nonlinear, dynamic and multivariable thermal process whose underlying mechanisms are not yet fully understood \cite{chasiotisArtificial2020}. It is a highly coupled multivariate problem considering the coupled heat, momentum, and mass transfer which, when modelled, can lead to insights into the underlying process and quality parameters. 

When only estimates of the MC are desired, a variety of soft-computing methods have proven effective for MC estimation during drying of different materials. 
A comparison of MC estimation performance using k-nearest neighbour regression, Support Vector Regression (SVR), Random Forest Regresssion (RFR), Artificial Neural Networks (ANN), and Gaussian processes, identified RFR as the most successful algorithm in estimation of drying characteristics and RFR and SVR as the most successful algorithms for MC estimation \cite{SaglamC_apple_slices}. 

MC estimation performance of basil seed mucilage using genetic algorithm-based ANNs and adaptive neuro-fuzzy inference systems (ANFIS) was studied by \cite{Amini2021}. Their results indicate that, while the ANFIS model gave the best total fit of MC, both the ANNs and ANFIS can give good estimations of MC during infra-red drying. 

Drying of marrow slices was investigated in \cite{Ceclu2022}. It was found that amongst the thin layer drying models, the Logarithmic, the Henderson, and Pabis models were the best. However, it was also found that a multi-layer perception (MLP) network using Backpropagation-based training was able to estimate the MC with an insignificant error. 
ANNs were also found successful in MC estimation of edible rose \cite{Qiu2022}, quince slices \cite{chasiotisArtificial2020}, green tea leaves \cite{kalathingal2020artificial}, absinthium leaves \cite{karimi2012optimization}, pistachios \cite{tavakolipour2012neural}, water melon rind pomace \cite{fabani2021producing}, and discarded yellow onions \cite{roman2020convective}. 

A mini-review by \cite{Yang2022} shows that ANNs, in general, are well suited for MC estimation applications utilising microwave drying, and \cite{aghbashlo_application_2015} shows that ANNs are well suited in general for foodstuff drying applications. 
ANNs have also shown to be a good tool for other estimation applications, such as estimating the State-of-Charge of batteries for electric vehicles \cite{lipu2019extreme,how2020state}, remaining useful lifetime of batteries \cite{9137406}, breaking pressure \cite{8119882}, solenoid valve remaining useful life-time \cite{9426406} and nitrogen in wheat leaves \cite{7742940}.

The above literature review shows that there exists a plethora of effective MC estimation techniques, which have been widely applied and researched in the field of foodstuff, especially for thin products (thickness magnitude approximately $10^{-4}$ meter to $10^{-2} $ meter). The filter media investigated in this work have a thickness magnitude of approximately $10^{-1}$ meter. It is therefore not a given that the results extend to this category of products. Furthermore, we have been unable to identify any studies of online MC estimation of filter media or similar products.

The objectives of this work are as follows:
\begin{enumerate}
    \item To present and share a dataset of industrial production drying data of bulky filter media drying.
    \item To device a method that can estimate the MC of filter media during the drying process to a degree that is useful for manufacturing.
    \item To compare, quantify and evaluate said method with the state-of-the-art MC estimation models found in the literature.
\end{enumerate}

The rest of this article is organised as follows: Section \ref{S:SectionII} describes the model selection process for determining the proposed ANN architecture for MC estimation. Section \ref{S:SectionIII} describes the experimental setup and data collection, including the dataset and software, and competing estimation models, used in the study. Section \ref{S:SectionIV} contains the results and discussions. Finally section \ref{S:SectionV} concludes the the article.
\section{Development of the artificial neural network}\label{S:SectionII}

We formulate MC estimation as a regression problem. Given the measured feature vector $\mathbf{x}$ as input, we use an ANN to map this multi-dimensional feature vector to the MC value at the output of the network. 
That is, the ANN acts as a parametric function $f(\mathbf{x},\mathbf{w})$, the parameters $\mathbf{w}$ of which are determined so as to minimize the loss function (\ref{eq:loss_func}) of the ANN's response w.r.t. to the real MC values measured on training data as described by
\begin{equation}\label{eq:loss_func}
	\mathcal{L} = \frac{1}{N}  \sum_{k=1}^N  \left(  MC_k - \widehat{MC}_k \right)^2,
\end{equation}
where $N$ is the total number of samples, $MC_k$ is the k'th experimentally measured moisture content, and $\widehat{MC}_k$ is the k'th estimate of the actual moisture content, i.e., it is the response of $f(\cdot,\mathbf{w})$ when the k'th sample in the training set is introduced to it.

The complexity of the function indicated by the structure of the ANN has an effect on the effectiveness of the method. We determine a good structure for the ANN by following the model selection process described next.

\subsection{Model Selection Process}
The estimation performance of an ANN depends on the amount of training data, the quality of training data, the chosen architecture of the neural network, the choice of hyper parameters such as activation functions, optimization algorithm, use of dropout and batch normalisation, choice of learning rate, and mini-batch size. Choosing the best architecture of a neural network is a research field in and of itself called Neural Architecture Search and multiple methods have been developed for this task \cite{ren2021comprehensive,kiranyaz2017pop,tran2020heterogeneous}.

The architecture of the neural network was chosen to be a feedforward multilayer perceptron (MLP) which is used for its simplicity and success in estimating MC of other drying experiments \cite{chasiotisArtificial2020,kalathingal_artificial_2020}. The Rectified Linear Unit (ReLU) \cite{NairH10} was used as activation function of the hidden neurons. The linear activation function was used for the output neuron enabling regression to all values on the real number line.
The architecture of the neural network, the learning rate and the mini-batch size were determined by performing model selection using the values shown in Table \ref{tbl:HyperParameterSearchOptions}. This was done by using the ASHA algorithm \cite{li2018massively} combined with the cross-validation process, and it was orchestrated using the \textit{Tune} framework \cite{liaw2018tune}. 
Table \ref{tbl:HyperParameterChosenOptions} shows the selected hyper parameters. For the rest of the hyper-parameters, such as optimizer, weight-decays, loss function, etc, we use values based on empirically established heuristics in previous works in the literature. For exact values see section \ref{sec:training_strategy}.

\begin{figure}
	\centering
	\includegraphics[width=3in]{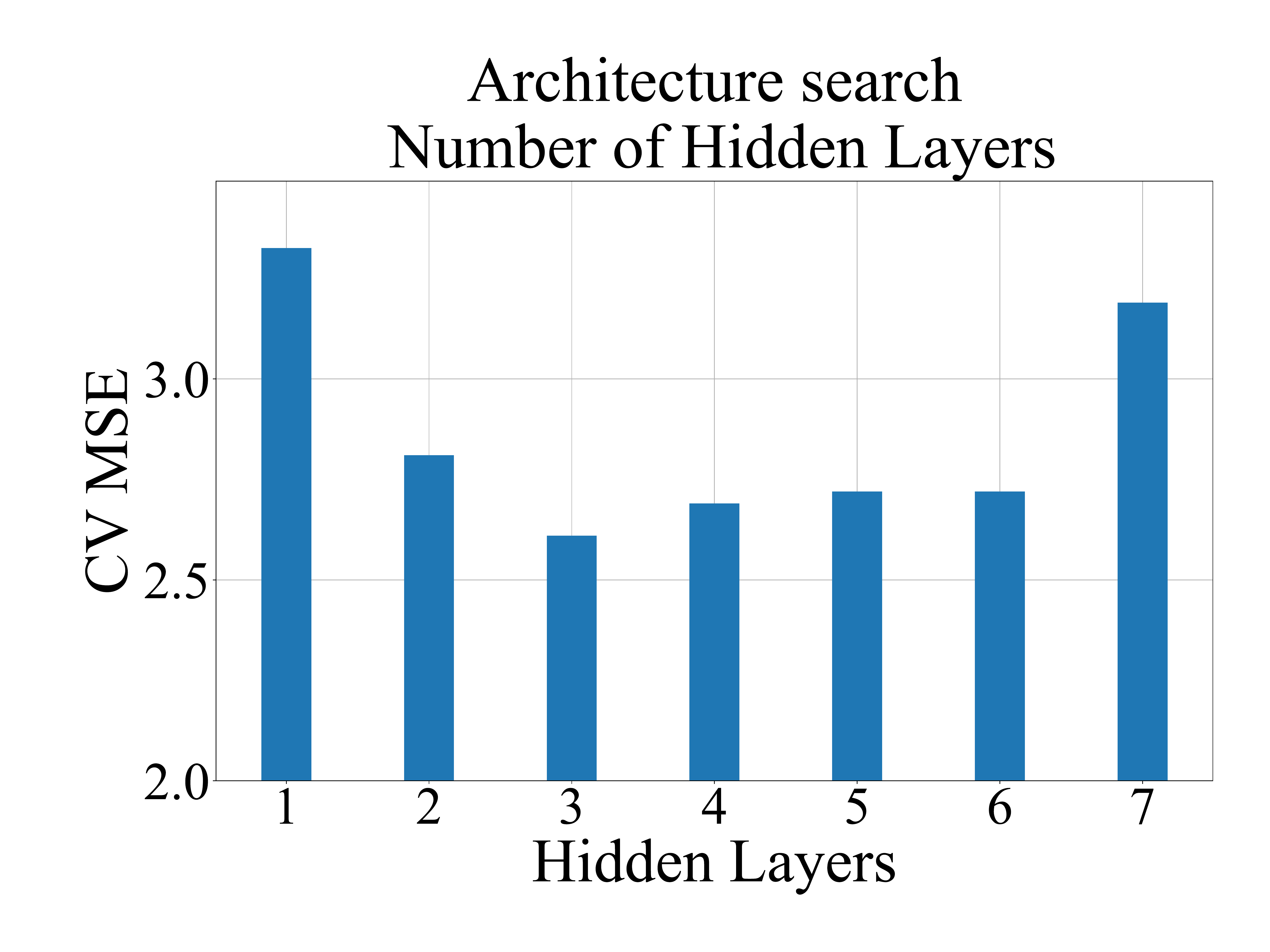}
	\caption{Average performance cross validation of found optimal hyper parameters for each size of the neural network based on 10 fold cross validation.}
	\label{fig-layer-search}
\end{figure}

\begin{figure}
	\centering
	\includegraphics[width=3in]{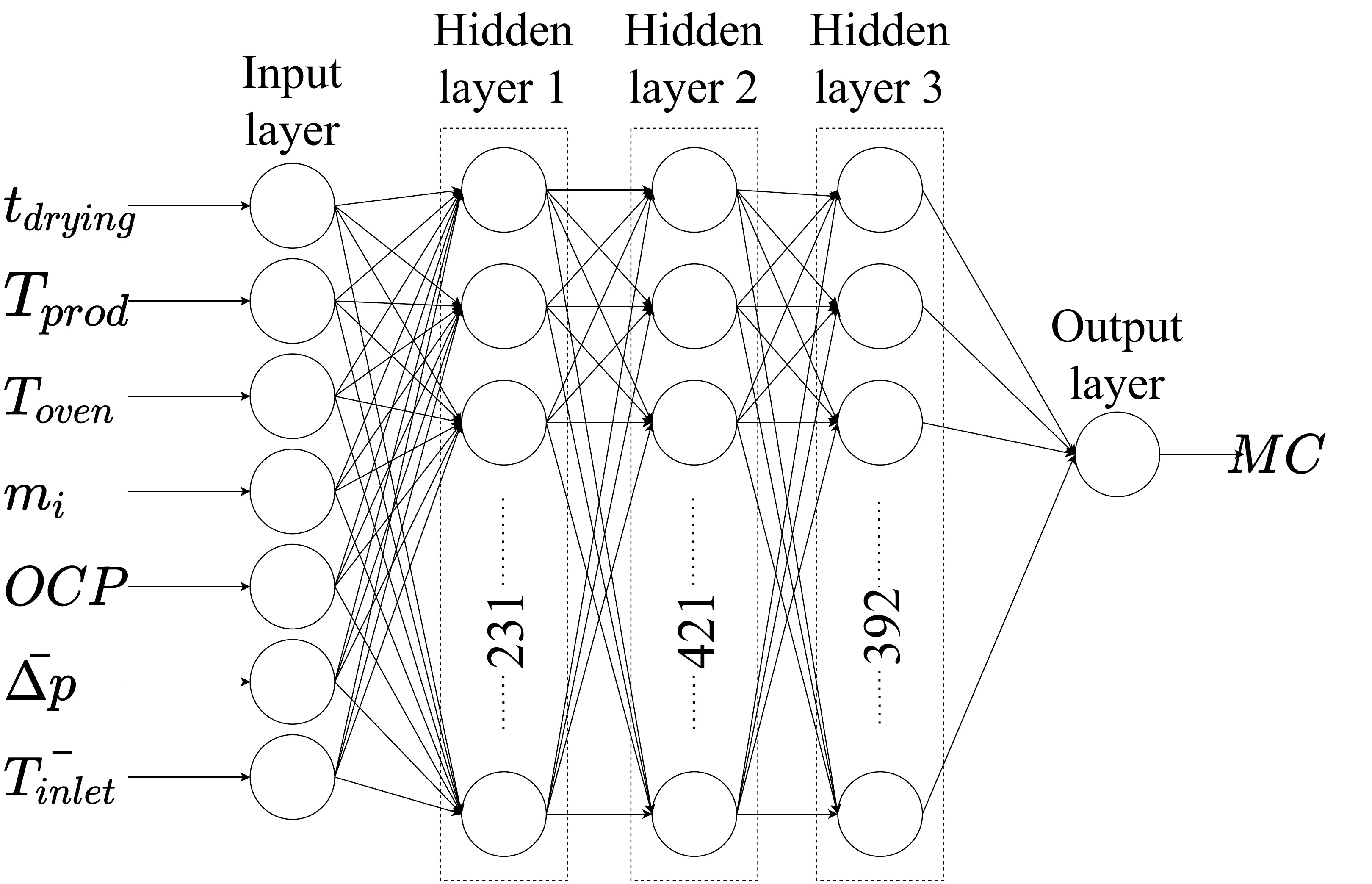}
	\caption{Schematic view of the found optimal neural network}
	\label{fig-optimal-nn-architecture-schematic}
\end{figure}

The expected error is highly dependent on the randomly chosen validation set, especially as the work in this article is based on a relatively sparse dataset with 322 sets of observations. To combat this issue, we combine cross-validation with the ASHA algorithm. However, instead of using the ASHA algorithm for early stopping of the training procedure, we test each set of chosen hyper parameters in a 10 fold cross validation loop. For each of the \textit{k} iterations, nine folds are used as training and validation (split 80/20) and the last fold is used for testing. The procedure is then repeated to 10 times. The ASHA algorithm is then applied to the mean of the mean squared error (MSE) losses of the test sets, with a grace period of three, a reduction factor of three, using one bracket. For each choice of number of hidden layers 500 trials were performed, the cross validation mean squared error was then plotted for each layer depth with its optimal parameters as can be seen in Fig. \ref{fig-layer-search}, resulting in a an optimal ANN architecture of which a schematic overview can be seen in Fig. \ref{fig-optimal-nn-architecture-schematic}.

\begin{table}[]
	\caption{Selection options for ANN hyperparameters}
	\label{tbl:HyperParameterSearchOptions}
	\begin{tabular}{lll}
		\hline \hline
		Hyperparameter 				& Type  		& Values \\ \hline
		Number of neurons 	        & continuous   	&   [1, 500]   \\
		Number of hidden layers  	& choice     	&   \{1, 2, 3, 4, 5, 6, 7\} \\
		Learning rate 				& continuous    &   log [1e-4, 1e-1] \\
		Batch size 					& choice		& \{2, 4, 8, 16, 32, 64\}  \\
		\hline \hline 
	\end{tabular}
\end{table}

\begin{table}[]
	\caption{Chosen configuration of hyper parameters.}
	\label{tbl:HyperParameterChosenOptions}
	\begin{tabular}{ll}
		\hline \hline
		Hyperparameter & Selected Parameters \\ \hline
		Number of neurons          &   $l_1=231$, $l_2=421$, $l_3=392$   \\
		Number of hidden layers          &    3  \\
		Learning rate          &   0.029924 \\
		Batch size & 16  \\
		\hline \hline 
	\end{tabular}
\end{table}

\subsection{Training Strategy} \label{sec:training_strategy}
Along with hyper parameter optimization, training of the parameters of the ANN is important in improving estimation quality. In order to find a converging point in the training process of the deep neural network, the Adam optimizer  \cite{kingma2014adam} was used with default decay rates of $\beta_1 = 0.9$ and $\beta_2 = 0.999$. The loss function used is the MSE of the MC estimates as defined by (\ref{eq:MSE}).

Batch normalisation was used before each layer to reduce the internal covariate shift of the activation functions  \cite{pmlr-v37-ioffe15}. Dropout \cite{srivastava2014dropout} was applied in the training phase after each hidden layer with a probability of attenuating each node of $50 \%$. Dropout can be seen as a form of data augmentation \cite{bouthillier2015dropout}, thus improving generalisation of the trained models.
In order to optimize training time, an early stopping scheme with a patience of 200 epochs was used, stopping the training when no improvements have been done on the validation set for 200 epochs. The best performing model was then saved and used for inference.

\subsection{Performance evaluation metrics}
The performance of the model is tested using four different measures, the MSE
\begin{equation}\label{eq:MSE}
	MSE =  \frac{1}{N} \sum_{k=1}^{N}  \left( MC_k-	\widehat{MC_k} \right)^2,
\end{equation}
the mean absolute error (MAE)
\begin{equation}\label{eq:MAE}
	MAE = \frac{1}{N} \sum_{k=1}^{N} \left( |  MC_k-	\widehat{MC_k}    | \right),
\end{equation}
which is less sensitive to outliers, as well as the standard deviation (STD)
\begin{equation}\label{eq:STD}
	STD = \frac{1}{N} \sum_{k=1}^{N} \left( |  MC_k-	\widehat{MC_k}    | \right),
\end{equation}
and the coefficient of determination $R^2$ between the estimates and the experimentally measured values
\begin{equation}\label{eq:R2}
	R^2 = 1 - \frac{\sum_{k=1}^{N} \left(MC_k - \widehat{MC_k}  \right) }{\sum_{k=1}^{N} \left(  MC_k - \widebar{MC} \right)}.
\end{equation}
where $\widebar{MC}$ is the average experimentally measured moisture content.

\section{Experimental setup and data collection}\label{S:SectionIII}

All drying experiments were conducted in a test oven concurrently drying four different filter media, replicating industrial usage. The positions in the oven are weakly coupled and each drying process can be approximated as an independent process. 

\subsection{Drying Procedure and Moisture Content}
The experimental MC was measured using the gravimetric method. 
The drying is split into two phases, namely Drying Phase 1 (DP1) and Drying Phase 2 (DP2). 
DP1 replicates the real-world industrial drying. However, in order to map out the entirety of the drying curve a variation in drying time is induced by extracting the filter media after a predetermined amount of time. 
DP2 lasts 48 hours with an oven temperature of $120 ^{\circ} C$. The purpose of DP2 is to evaporate all MC from the filter media thus enabling the measurement of the solid mass $m_{solid}$ which is used to calculate the experimental MC in the filter media, as seen in (\ref{eq_moisture-content_initial}) and (\ref{eq_moisture-content}).

The mass of the filter media are measured three times during the experiment. The initial (wet) mass, $m_{initial}$, of each filter media is measured before DP1.
$m_{after}$ is measured after DP1, and $m_{solid}$ which is measured after DP2. 

With these measured masses we can now calculate the initial MC as:
\begin{equation} \label{eq_moisture-content_initial}
	MC_{initial} = \frac{m_{initial}-m_{solid}}{m_{solid}}  100 \%,
\end{equation}
and the MC after DP1 as:
\begin{equation} \label{eq_moisture-content}
	MC = \frac{m_{after}-m_{solid}}{m_{solid}}  100 \%.
\end{equation}

\subsection{Dataset}

\begin{figure*}
\centering
\subfloat[]{\includegraphics[height=2.5in]{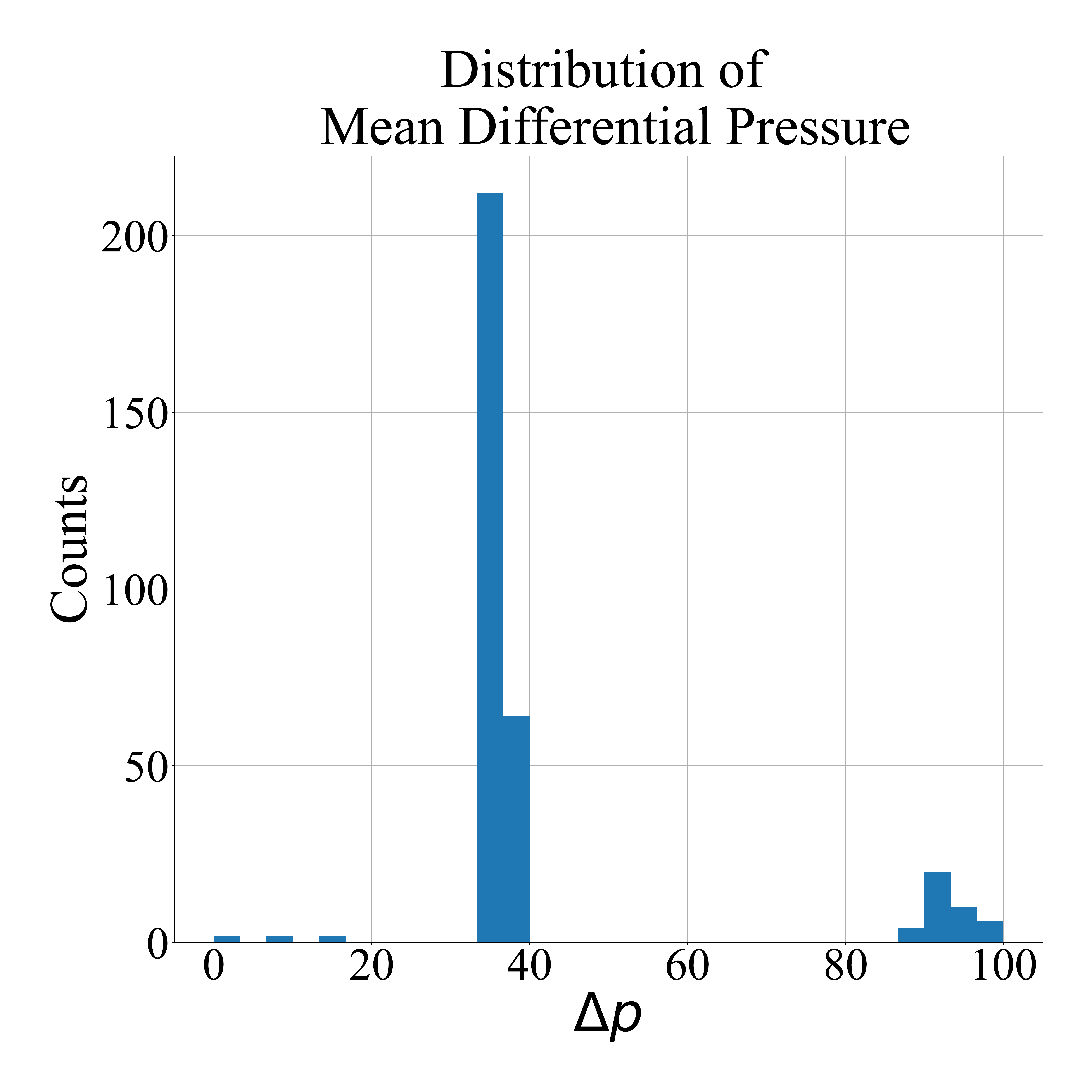}\label{fig-dataset-distribution-mean_oven_dp}} %
\subfloat[]{\includegraphics[height=2.5in]{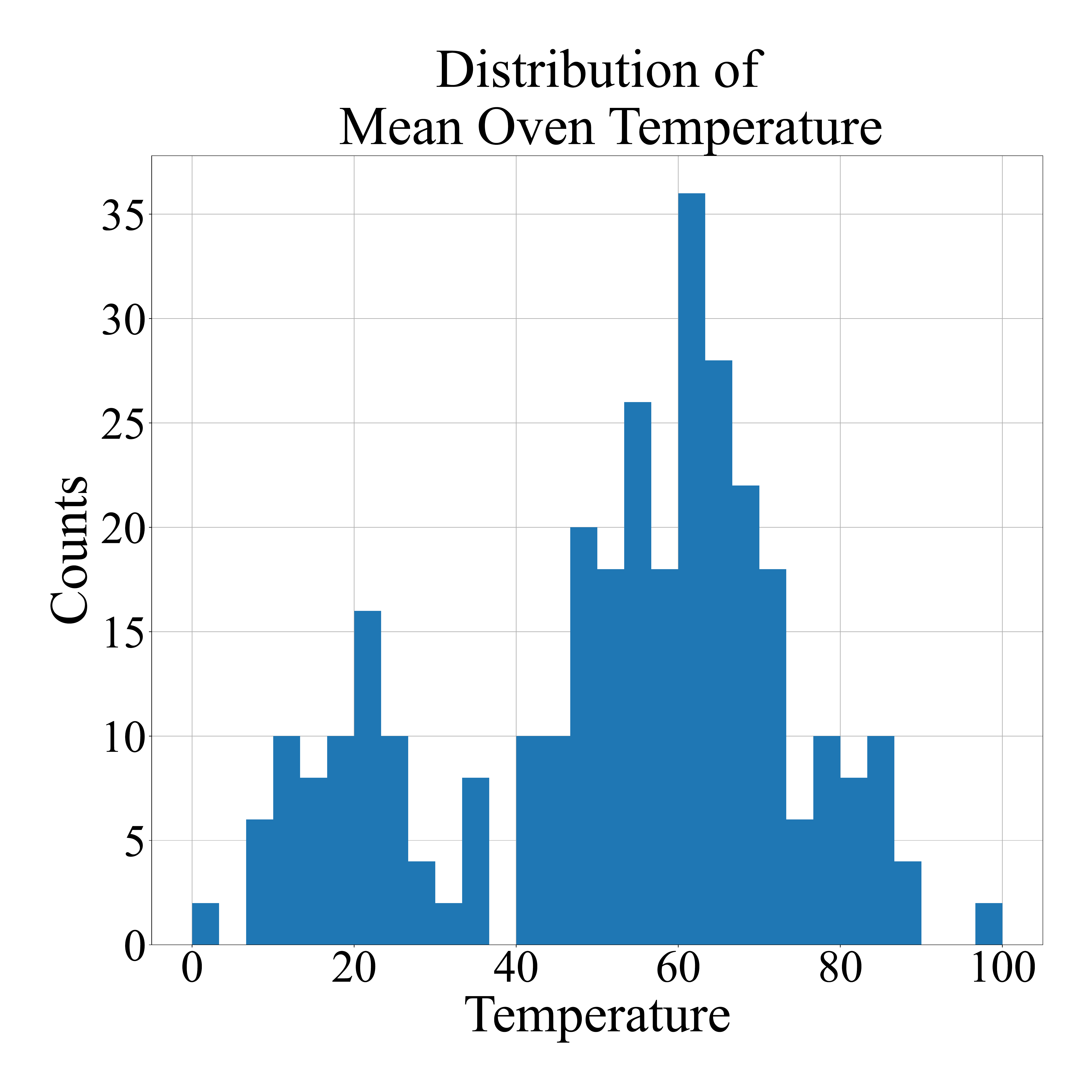}\label{fig-dataset-distribution-mean_oven_temperature}} %
\subfloat[]{\includegraphics[height=2.5in]{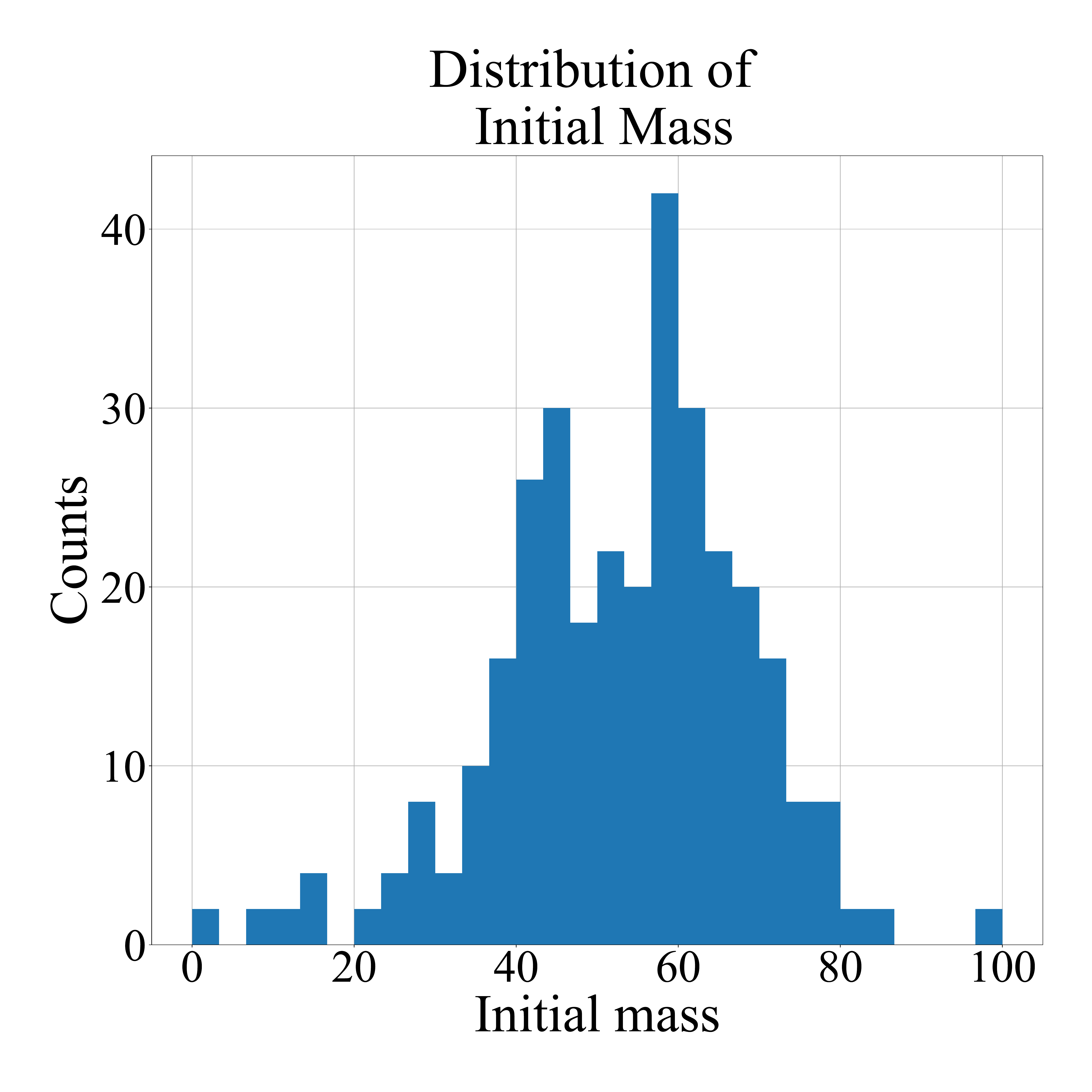}\label{fig-dataset-distribution-initialmass}} %
\\
\subfloat[]{\includegraphics[height=2.5in]{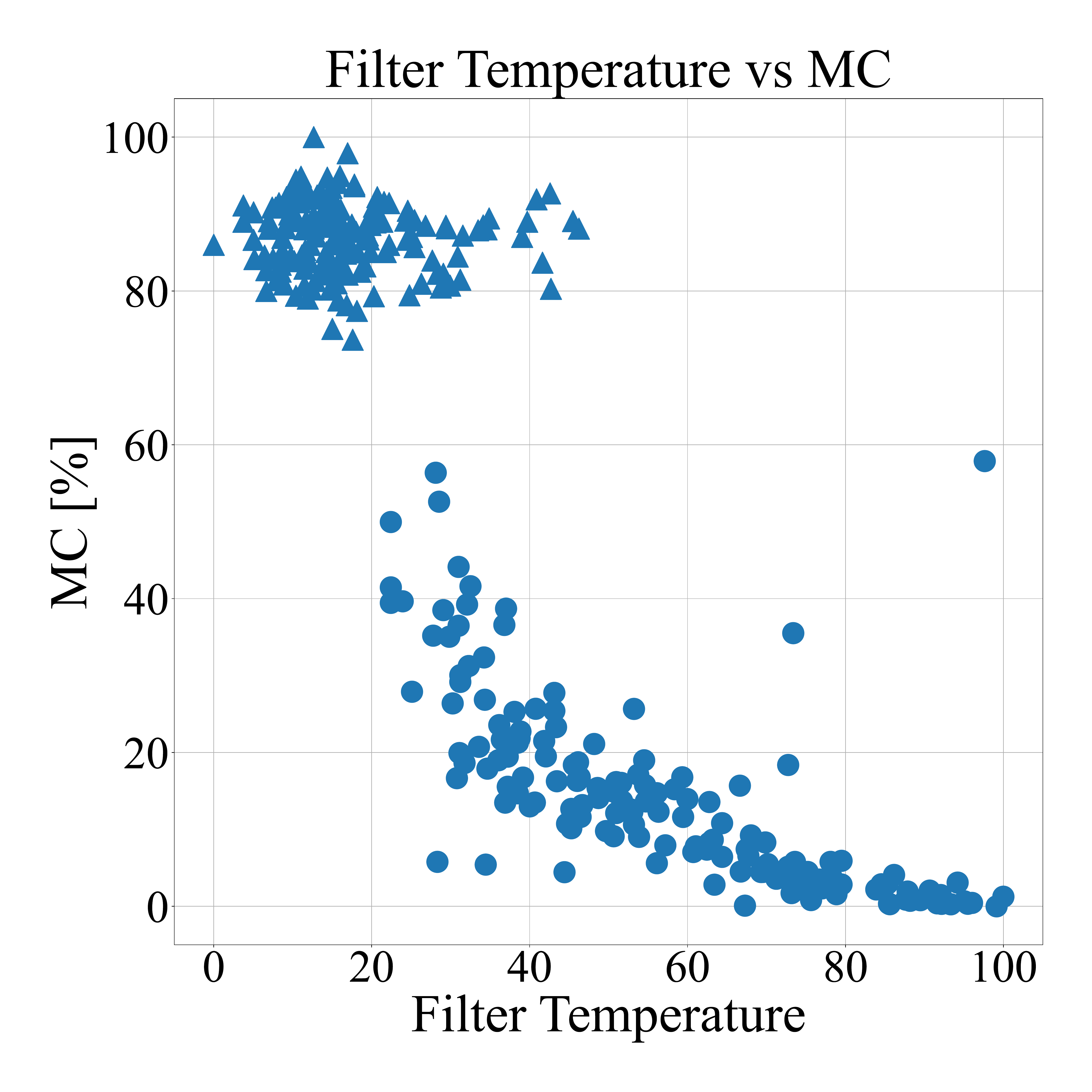}	\label{fig-dataset-temperature-vs-MC}}
\subfloat[]{\includegraphics[height=2.5in]{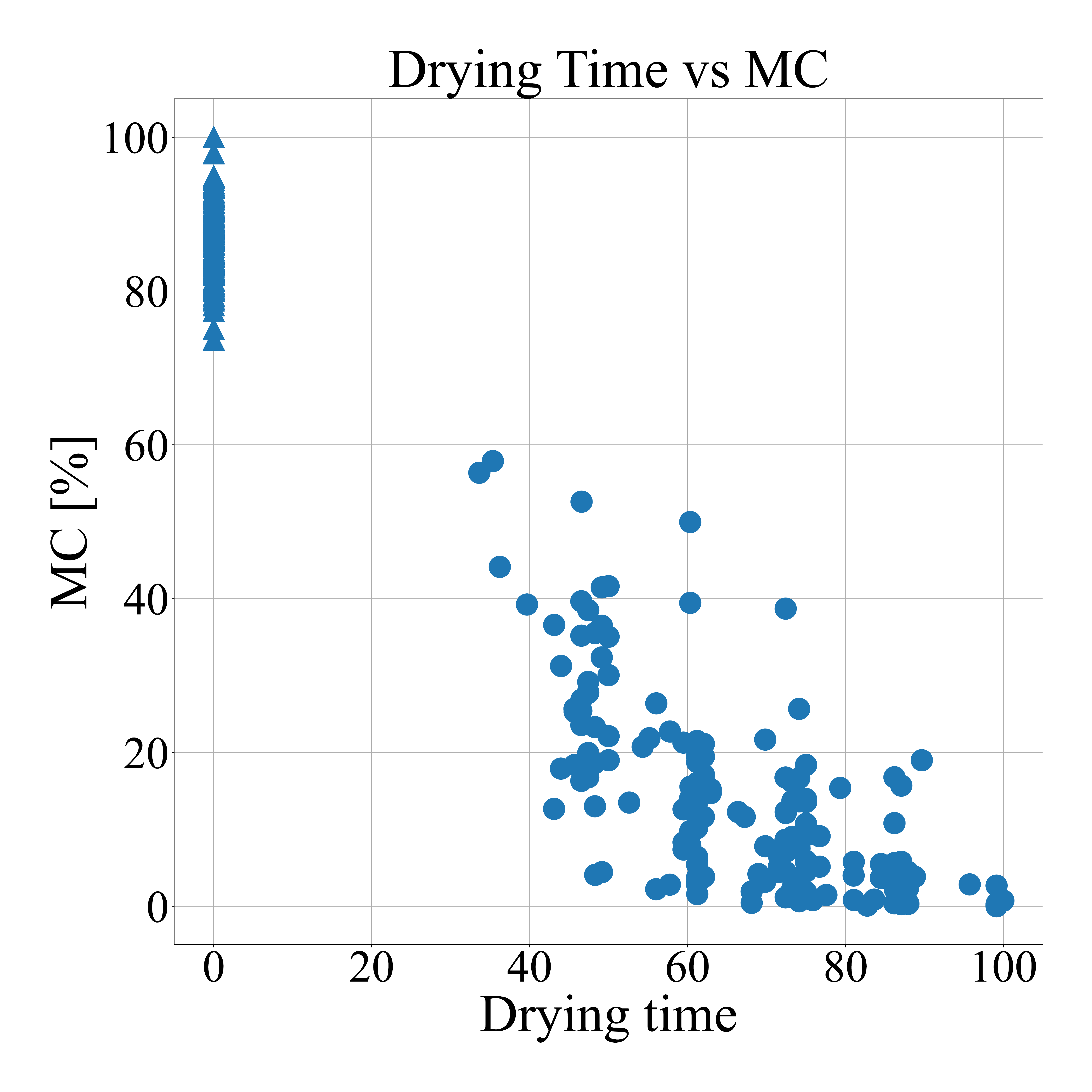}	\label{fig-dataset-dryingtime-vs-MC}}
\caption{ (a) Distribution of normalized dimensionless mean differential pressure of each filter media during the drying process. (b) Distribution of the normalized dimensionless mean oven temperature of each drying experiment. (c) Distribution of the normalized dimensionless initial mass of filter media. (d). Normalized dimensionless estimated filter media temperature at extraction time as a function of MC. A clear relationship between estimated filter media temperature and MC can be seen. Cluster in upper left corner corresponds to the ICD. (e) MC as a function of normalized dimensionless drying time. Large variance in both drying time and MC can be seen. Cluster in upper left corner corresponds to the ICD.}
\end{figure*}

Automated data collection is used to collect the seven predictor variables that constitute the dataset. The drying time $t_{drying}$, the estimated filter media temperature $\widehat{T}_{filter}$, the oven chamber position $OCP$, the overall mean of the oven input temperature during the drying process of the particular filter media $\widebar{T}_{in}$, the overall mean of the differential pressure across the oven $\widebar{\Delta p}$, the oven temperature at the time of filter extraction $T_{cur}$, and the initial mass of the filter media before drying $m_i$ are collected for each experiment.

Each measurement of the dataset was normalized and scaled such that all values lie in the range of $[0,100]$. The normalized values of a sample $\mathbf{x}$ were calculated using:
\begin{equation}
	\mathbf{z}_k =  \frac{\mathbf{x}_k - min(\mathbf{x}_k^{train})}{max(\mathbf{x}_k^{train})-min(\mathbf{x}_k^{train})} \cdot 100,
\end{equation}
where $\mathbf{z}_k$ is the vector of normalized values of feature $k$, $\mathbf{x}_k$ is the vector of all values of feature $k$, $\mathbf{x}_k^{train}$ is the vector of values of feature $k$ belonging to the training set.

A total of 161 experiments were performed resulting in 322 sets of predictor- and response variable vectors. 161 sets of observations measuring the \textit{initial condition data} (ICD) and 161 sets of predictor- and response variable observations with different drying times. The dataset consists of two classes of datapoints, ICD and \textit{end condition data} (ECD), where the ICD are the sets of observation sampled upon insertion of a filter media into the drying oven, i.e. a drying time of zero minutes. The ICD are information poor, as an equilibrium has not been reached yet and, as an effect, the sensors are sensing the features of the oven and not those of the filter media. The ECD are the sets of observations upon extraction of the filter media from the drying oven, i.e. after the designated drying time for the specific filter media. The ECD are relatively information rich, and regression or estimation can be utilized. The dataset is published in the IEEE DataPort repository and can be found here: https://dx.doi.org/10.21227/hwa2-tp66 \cite{hwa2-tp66-22}.

The features of the dataset can furthermore be classified into two feature types, i.e., status features and oven setting features.

\subsubsection{Oven setting features}
The oven setting features are the features describing the physical environment in which the filter is dried. The oven setting features are the position of the filter media in the oven, the mean oven temperature during the drying time of each specific filter media, the mean oven differential pressure during the specific drying time of the filter media, the current oven temperature, and the initial mass of the filter media before drying begins.

Fig. \ref{fig-dataset-distribution-mean_oven_dp} shows the distribution of the differential pressure over the fan pushing the air into the oven. The differential pressure is correlated with air speed, and thus the mass of air circulating in the oven. As can be seen, one set of 20 drying experiments has been done under other circumstances than the rest of the filter media, and a trained model will need to be able to encompass this deviation in oven setting parameters as well. The outlier data has been included as it will serve to challenge the performance of the produced models.

Fig. \ref{fig-dataset-distribution-mean_oven_temperature} shows the distribution of the mean oven temperature during the drying experiments of each filter media. Here, a binormal distribution can be seen. This is due to the unfortunate deconstruction and reconstruction of the test oven during the multi-month data acquisition period. If the estimation models are able to encompass these different oven setting features, then it only bodes well for the generalizability of the model.

Fig. \ref{fig-dataset-distribution-initialmass} shows the distribution of the initial mass which as can be seen follows a skewed Gaussian distribution.

\subsubsection{Status features}
The status features are the features correlated with the current drying status of the filter i.e., the drying time and the estimated filter media temperature.

Fig. \ref{fig-dataset-temperature-vs-MC} shows the MC as a function of the dimensionless normalised estimated filter media temperature. A clearly dependent relationship between the MC and the estimated filter media temperature can be identified, the lower the temperature the larger the variation in MC as is expected from the behaviour of a typical drying curve. The estimated filter media temperature holds much of the information that the proposed models will be able to utilize in order to make good estimates.

Fig. \ref{fig-dataset-dryingtime-vs-MC} shows a relationship between drying time and MC. There is a large variance along the MC axis, especially for lower drying times. This variance is where the possible gains of utilizing MC estimation can be seen. All low-drying-time or low-MC datapoints represents the possible optimization gains, as early stopping of the drying process can be done if identification of the MC is possible.

\subsection{Competing Estimation Models}
The proposed ANN-based approach is compared with data-driven models reported as state of the art for different MC estimation applications in the literature. To establish a baseline performance we use semi-empirical thin layer drying models, see Table \ref{tbl:methodology:thin_layer_models}. The thin layer drying models are all fitted using nonlinear least squares in the Matlab curve fitting toolbox \cite{matlabcurvefitting}. 

\begin{table}[]
    \caption{Thin-layer drying models}
    \label{tbl:methodology:thin_layer_models}
    \begin{tabular}{lll}
    Model            & Equation                    & Reference \\ \hline
    Lewis            & $MC =\exp(-kt)$                           &  \cite{lewis1921rate}         \\ 
    Page             & $MC = \exp(-kt^n)$                        &   \cite{page1949factors}        \\ 
    Two term        & $MC = a\exp(-k_1t)+b\cdot\exp(-k_2t)$      &  \cite{madamba1996thin}         \\ 
    Henderson       & $MC = a\exp(-kt)$                         &   \cite{hendersonPabis}        \\ 
    Logarithmic     & $MC = a\exp(-kt)+c$                       &  \cite{yaugciouglu1999drying}         \\ 
    Midilli et al.  & $MC = a\exp(-kt^n)+bt$                    &  \cite{midilli2002new}    \\ \hline
    \end{tabular}
\end{table}

Furthermore, we compare the ANN approach to SVR and RFR as reported by \cite{SaglamC_apple_slices}, and ANFIS as reported by \cite{Amini2021}, and partial least squares (PLS) to act as a baseline for the machine learning models. 
All competing models estimate the MC as output. The input for the thin layer drying models is solely drying time. The input for the machine learning models are all the same as that for the ANN. 

All models come in two variations. One trained on the entirety of the data, referred to as With Initial Conditions (WIC), and one trained only on the ECD, referred to as No Initial Conditions (NIC). As postulated earlier, the ICD is relatively information poor, and thus might hamper the estimation performance in the range of interest, the ECD. For practical applications, the quality of the estimates on the ICD can be ignored - as it is a trivial case.

\subsection{Model Performance Validation}
All models are validated using repeated 10 fold cross validation as described by \cite{Burman1989}. Regular 10 fold cross validation was performed by splitting the data into 10 folds, training on all but one fold, and then using the left-out fold for validation. This process was then repeated across all 10 folds, resulting in averages of the estimation error measures as described in (\ref{eq:MSE}), (\ref{eq:MAE}), (\ref{eq:STD}), and (\ref{eq:R2}). 
The data was then shuffled, and the above process was repeated five times. Therefore, all results reported in this section are based on validation data and not training data. Furthermore, all results reported are averages of the five times repeated 10 fold cross validation trials.

\section{Results and discussion}\label{S:SectionIV}

\subsection{Model Estimation performance}

\subsubsection{Thin Layer Drying models}
\begin{figure*}[]
    \vspace{-2.5cm}
    \centering
    \subfloat[]{\includegraphics[width=2in]{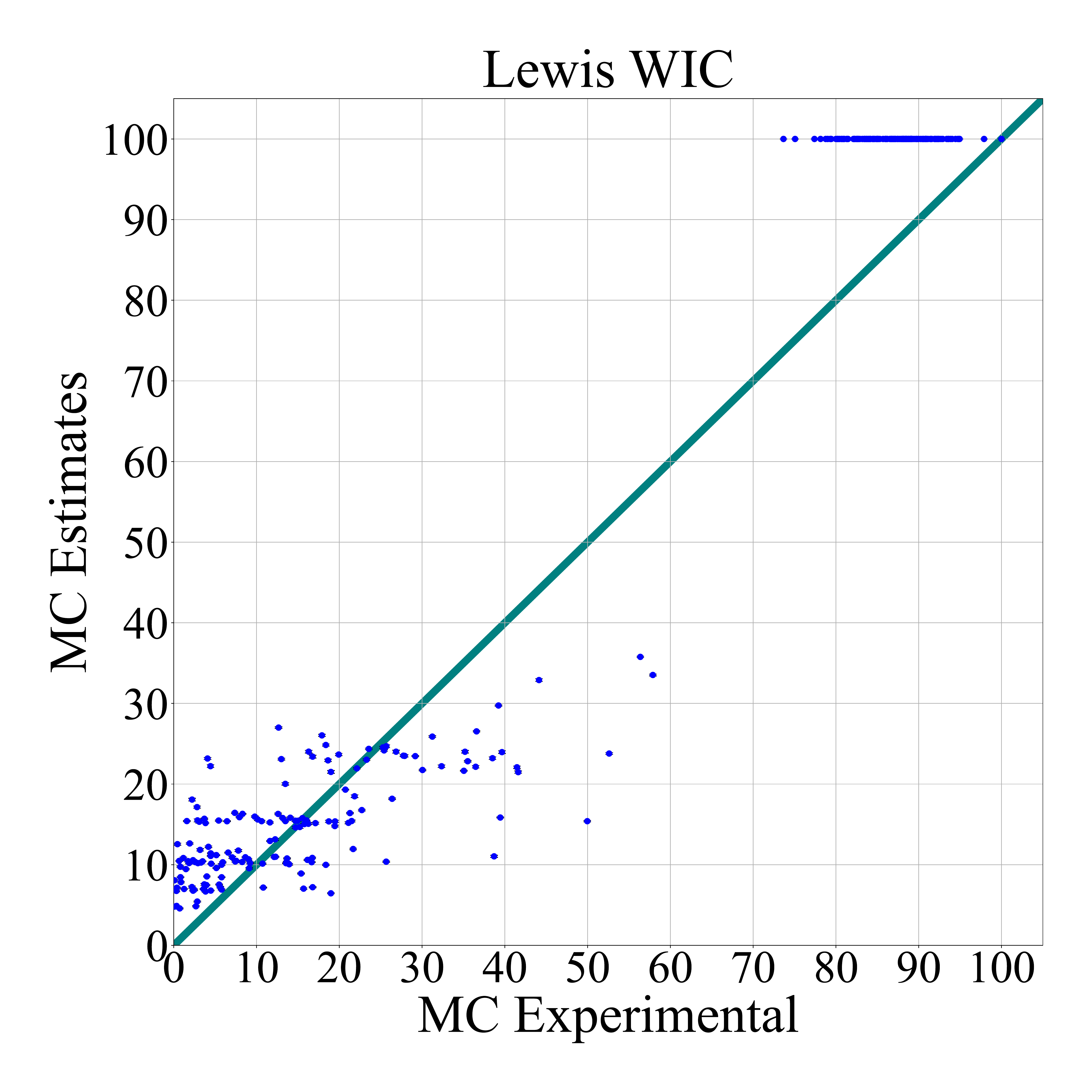} \label{fig-results-newton_WIC_preds}}  \,
    \subfloat[]{\includegraphics[width=2in]{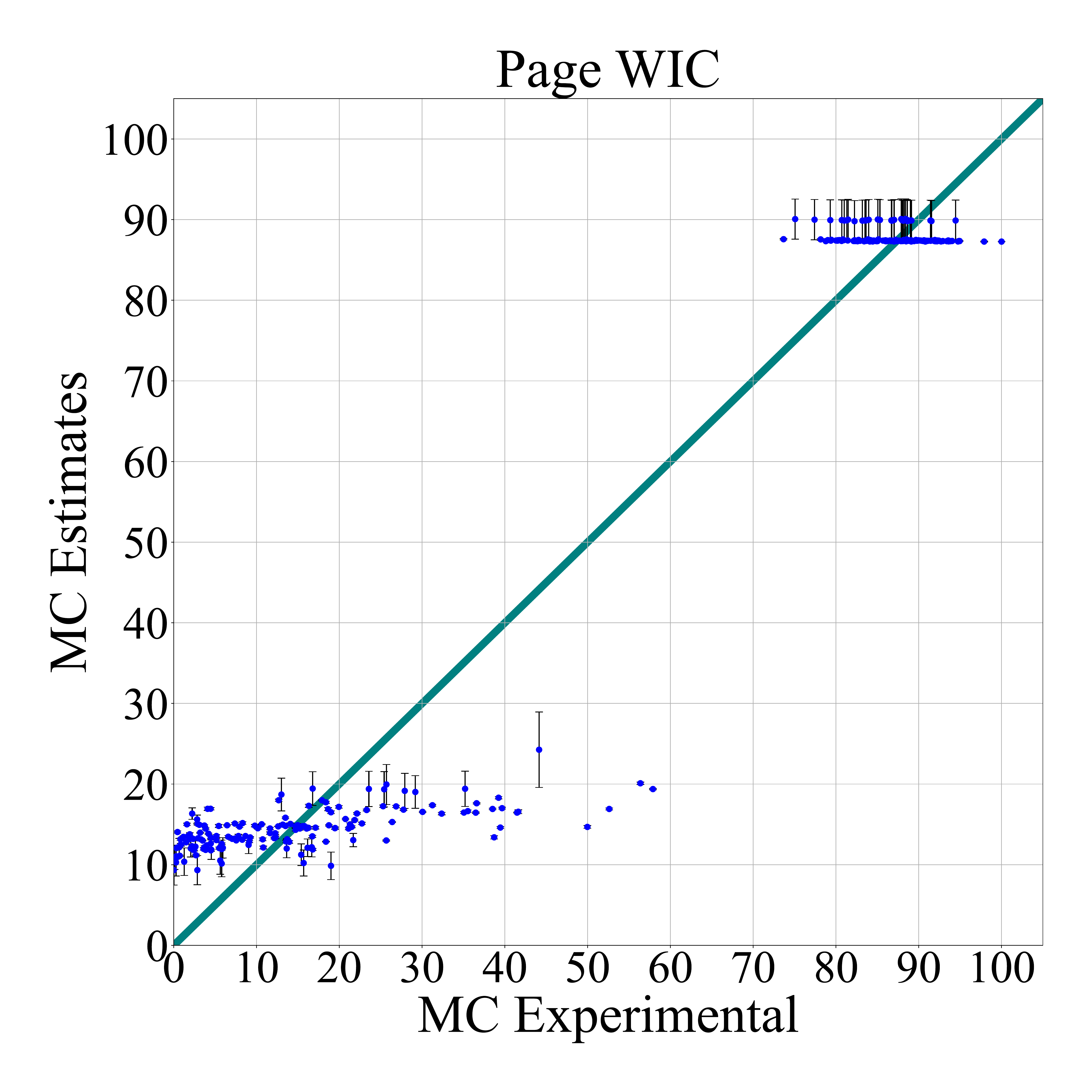} \label{fig-results-modified_page_WIC_preds}} \,
    \subfloat[]{\includegraphics[width=2in]{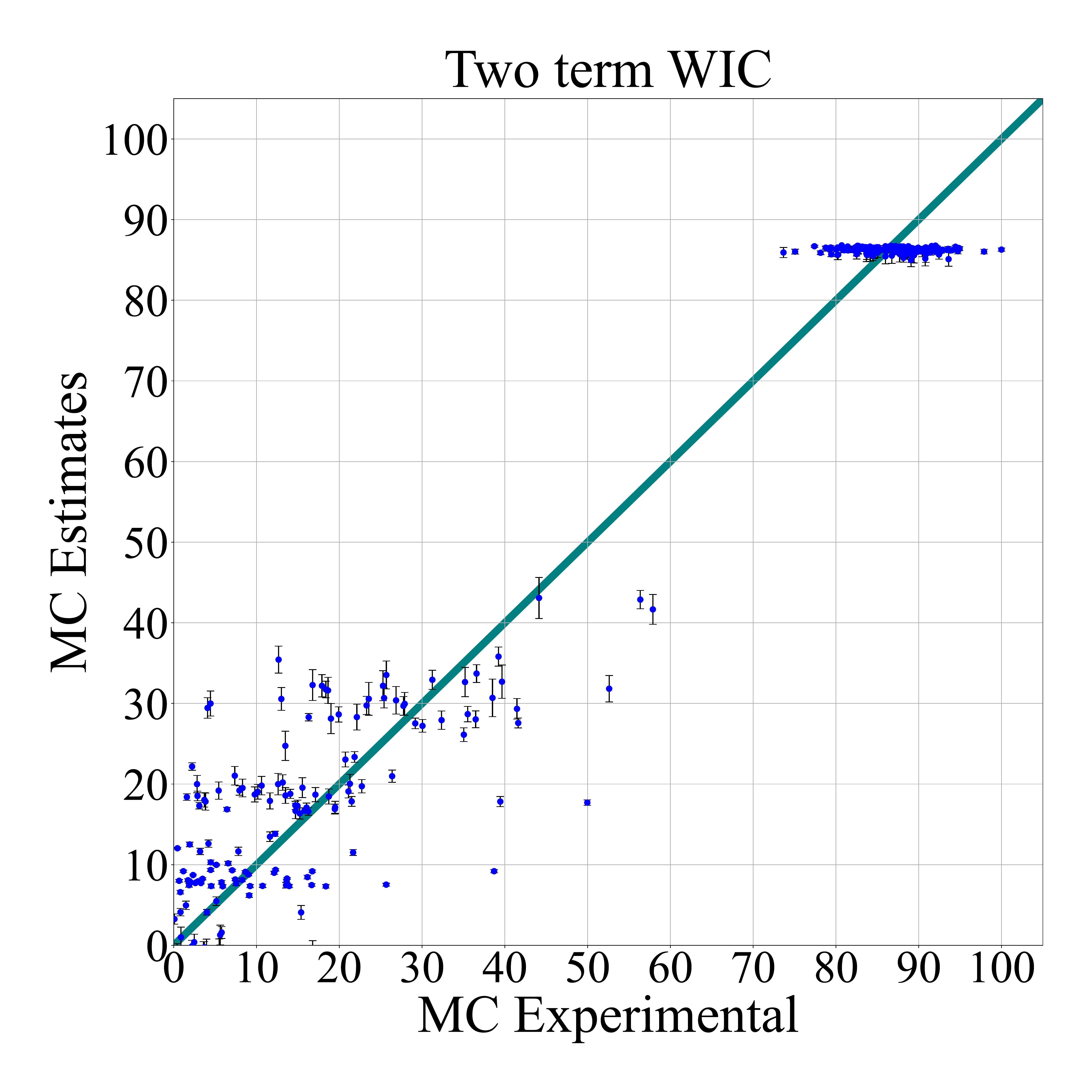} \label{fig-results-two_term_WIC_preds}}  
    \\[-0.2ex]
    \subfloat[]{\includegraphics[width=2in]{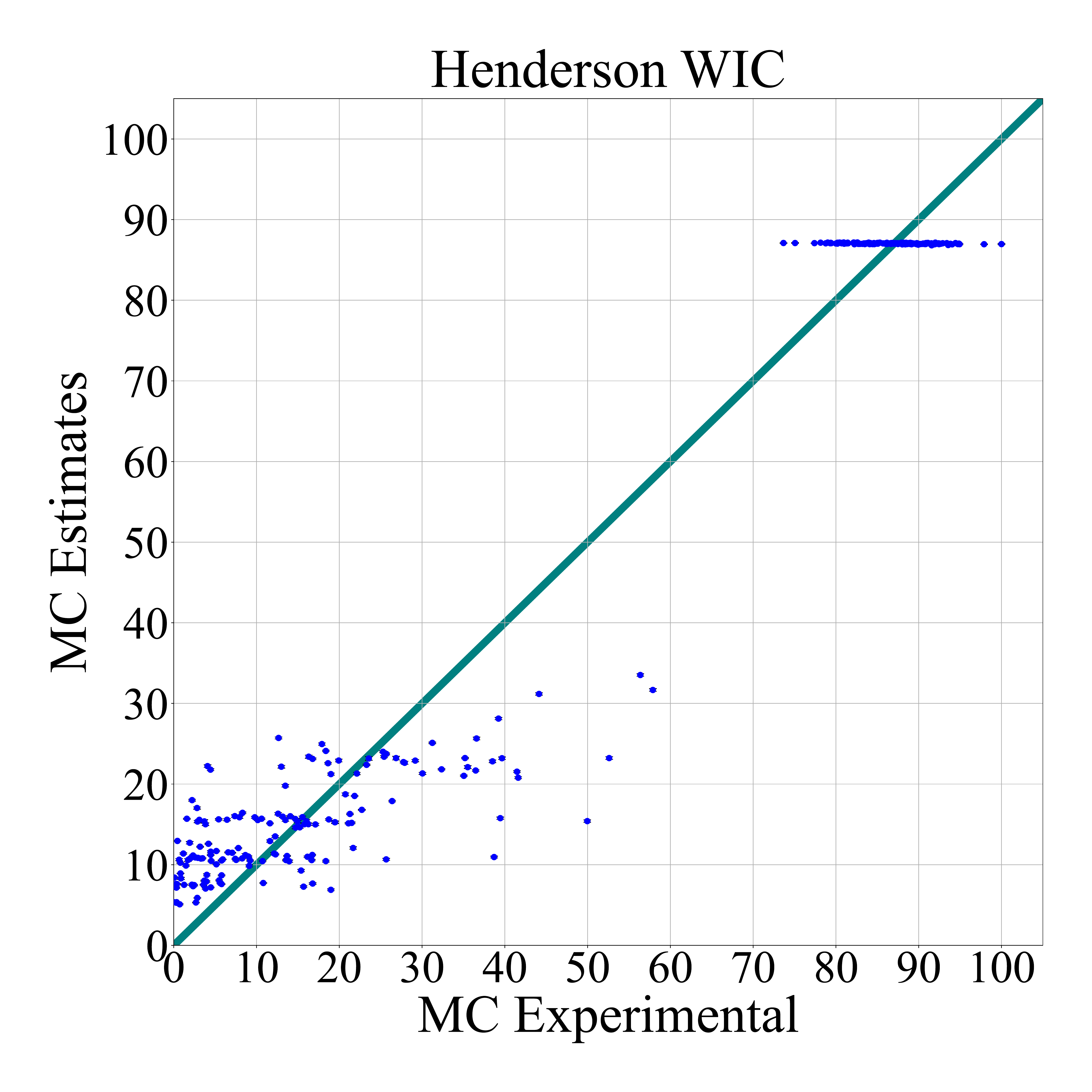} \label{fig-results-henderson_WIC_preds}} \,
    \subfloat[]{\includegraphics[width=2in]{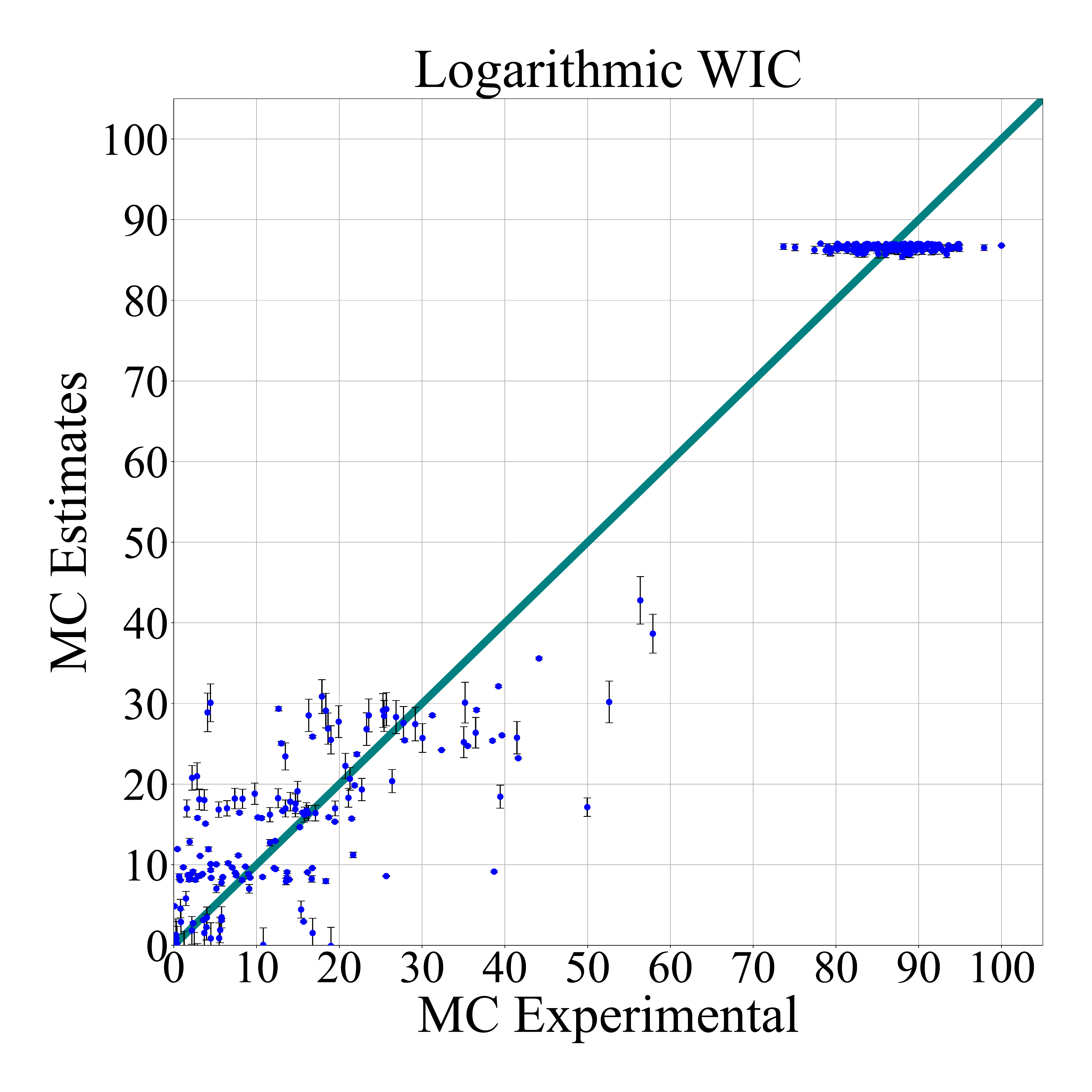} \label{fig-results-logarithmic_WIC_preds}} \,
    \subfloat[]{\includegraphics[width=2in]{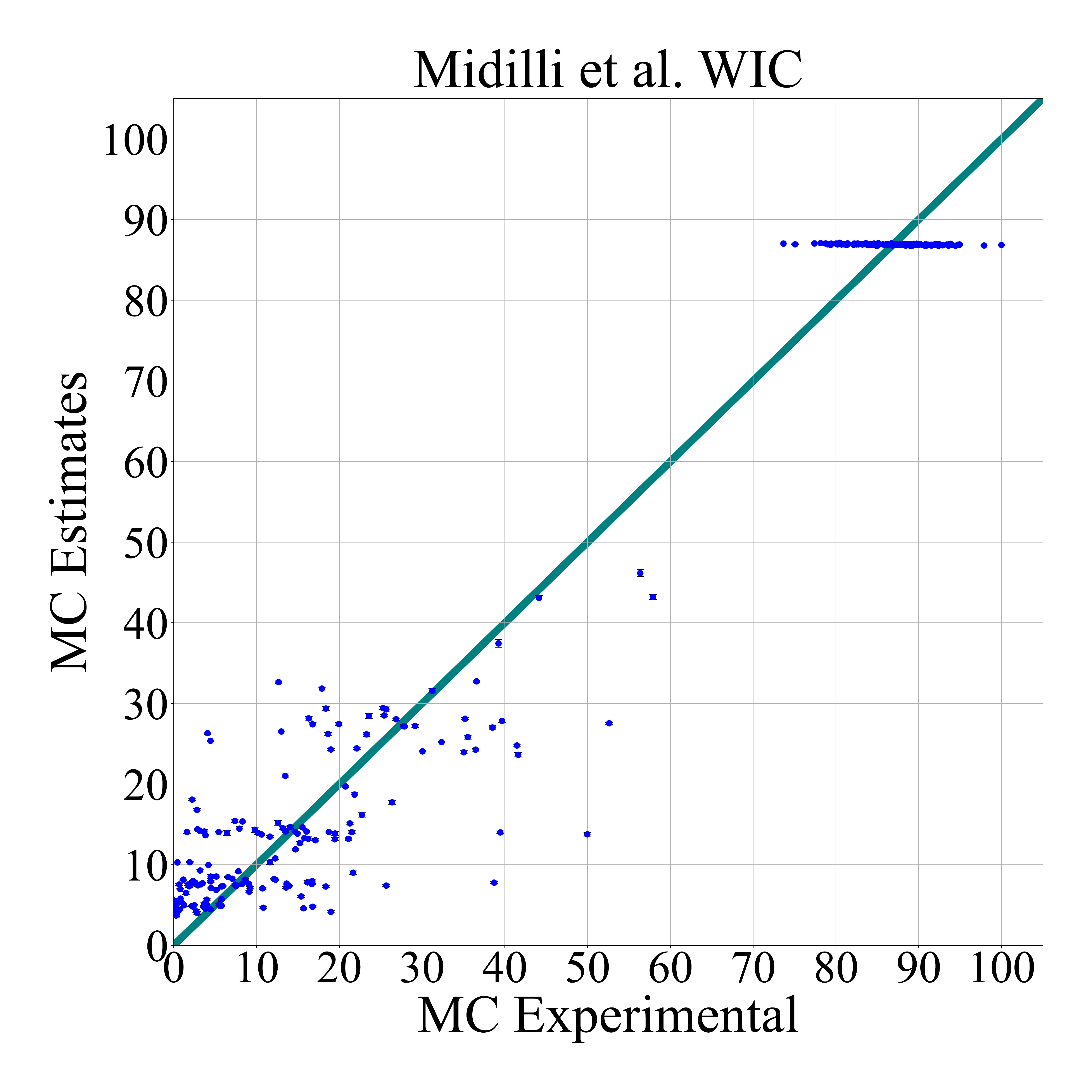} \label{fig-results-midilli_WIC_preds}} 
    \caption{Test fold results of model MC estimates for semi-empirical thin-layer drying models fitted using both ECD and ICD. Blue line indicates perfect estimation. Error bars represent the standard error of the mean, for the five times repeated cross-validation trials.}
    \label{fig:Thin_layer_drying_predictions_WIC}
    \vspace{-2.5cm}
\end{figure*}

\begin{figure*}[]
    \centering
    \subfloat[]{\includegraphics[width=2in]{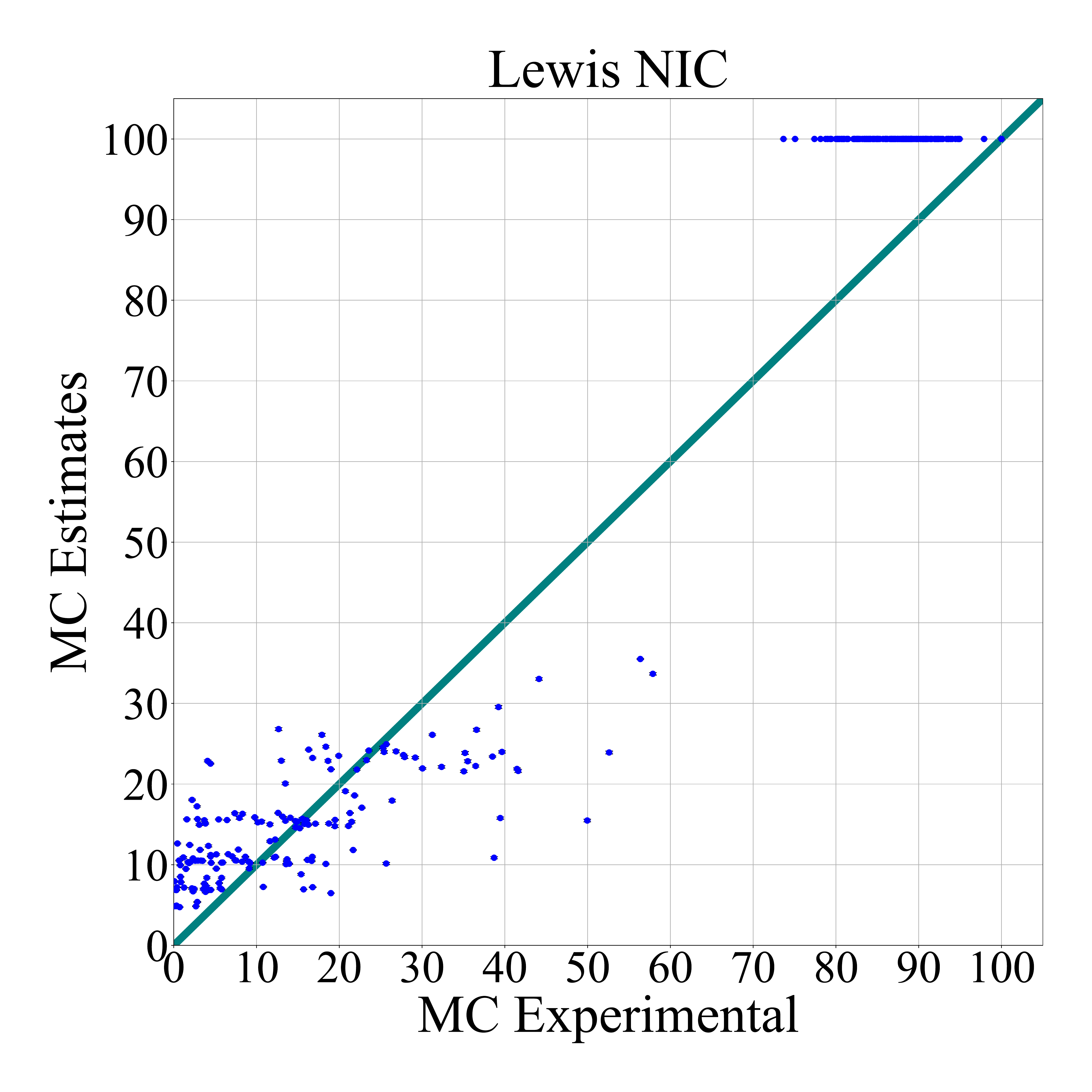} \label{fig-results-newton_NIC_preds}}  \,
    \subfloat[]{\includegraphics[width=2in]{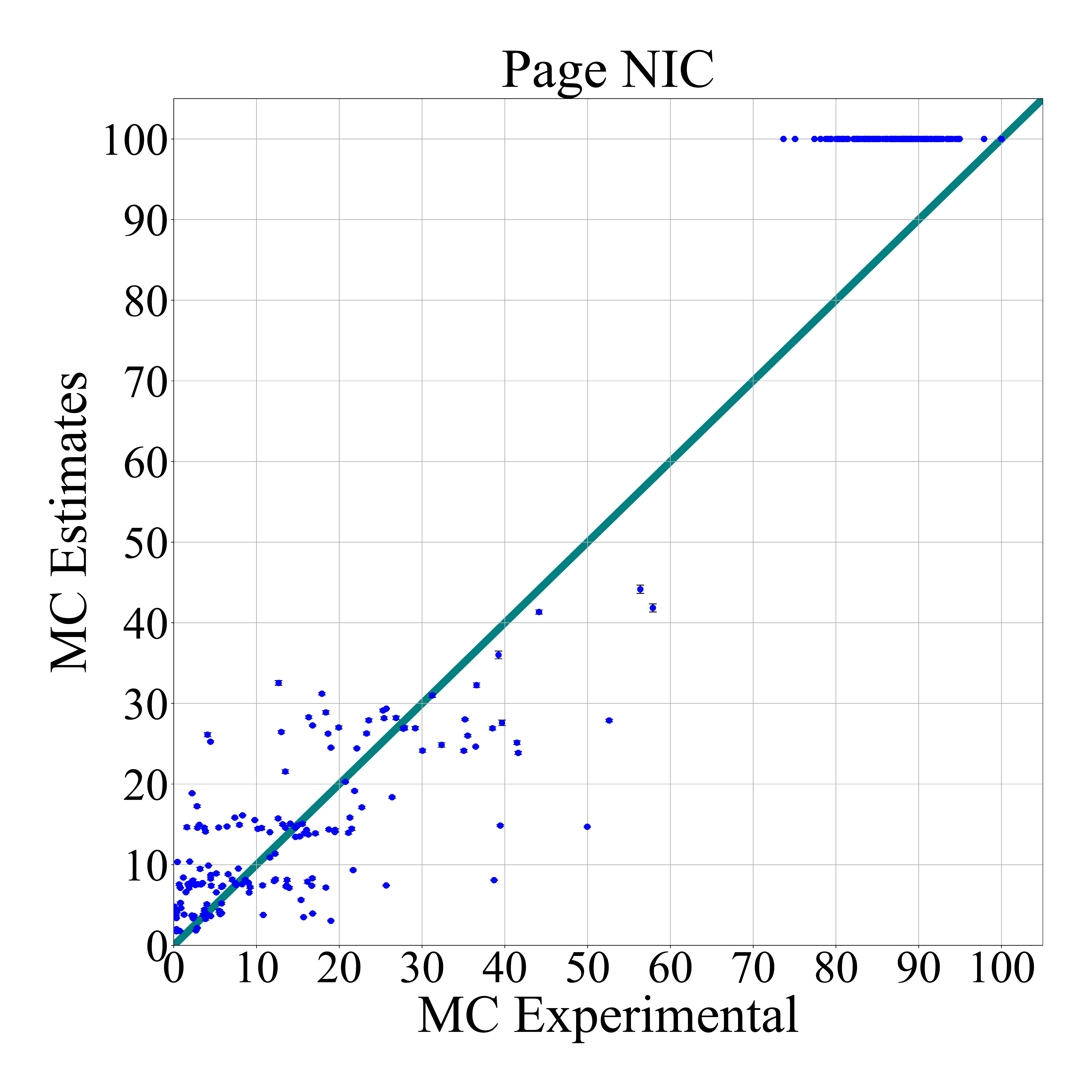} \label{fig-results-modified_page_NIC_preds}} \,
    \subfloat[]{\includegraphics[width=2in]{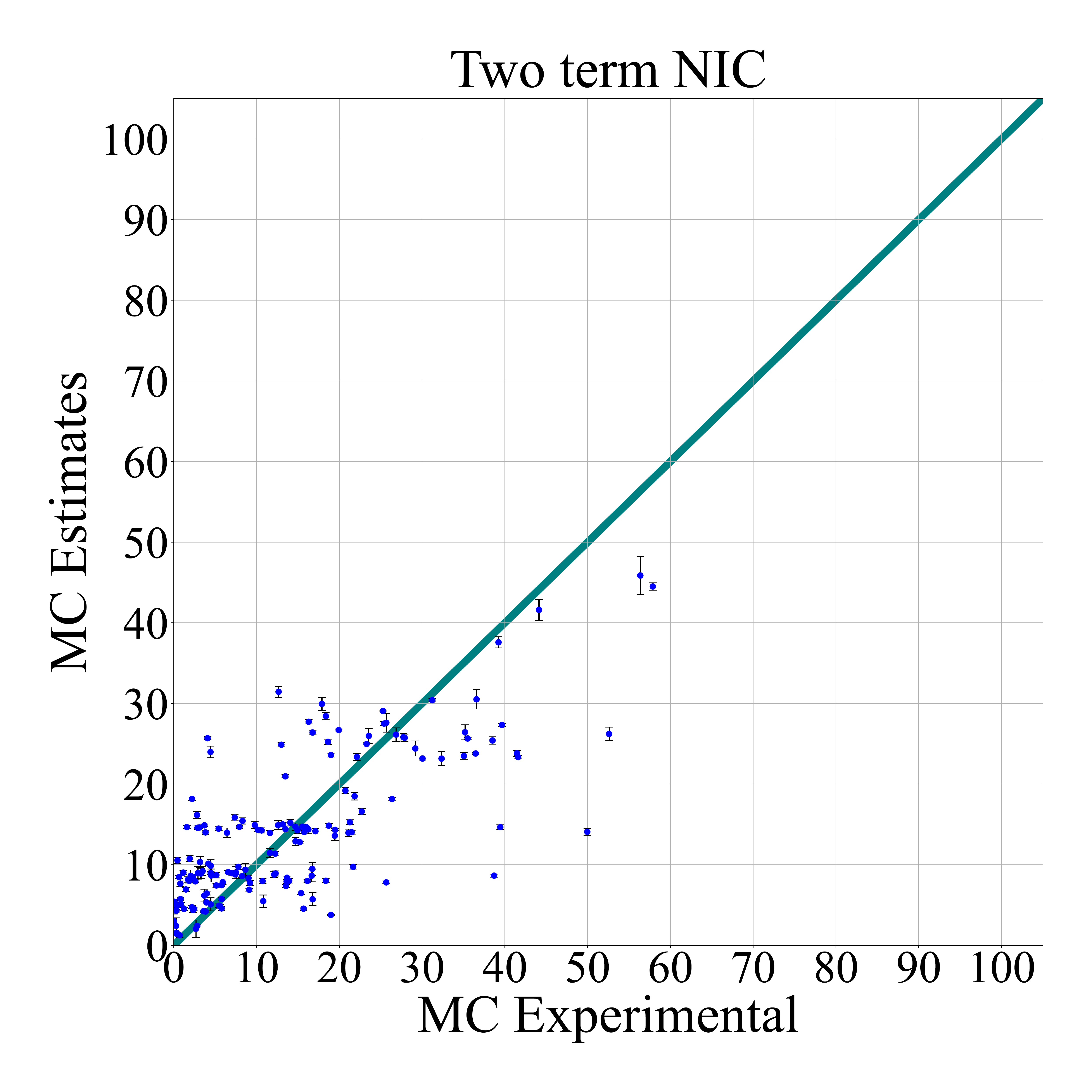} \label{fig-results-two_term_NIC_preds}}  
    \\[-0.2ex]
    \subfloat[]{\includegraphics[width=2in]{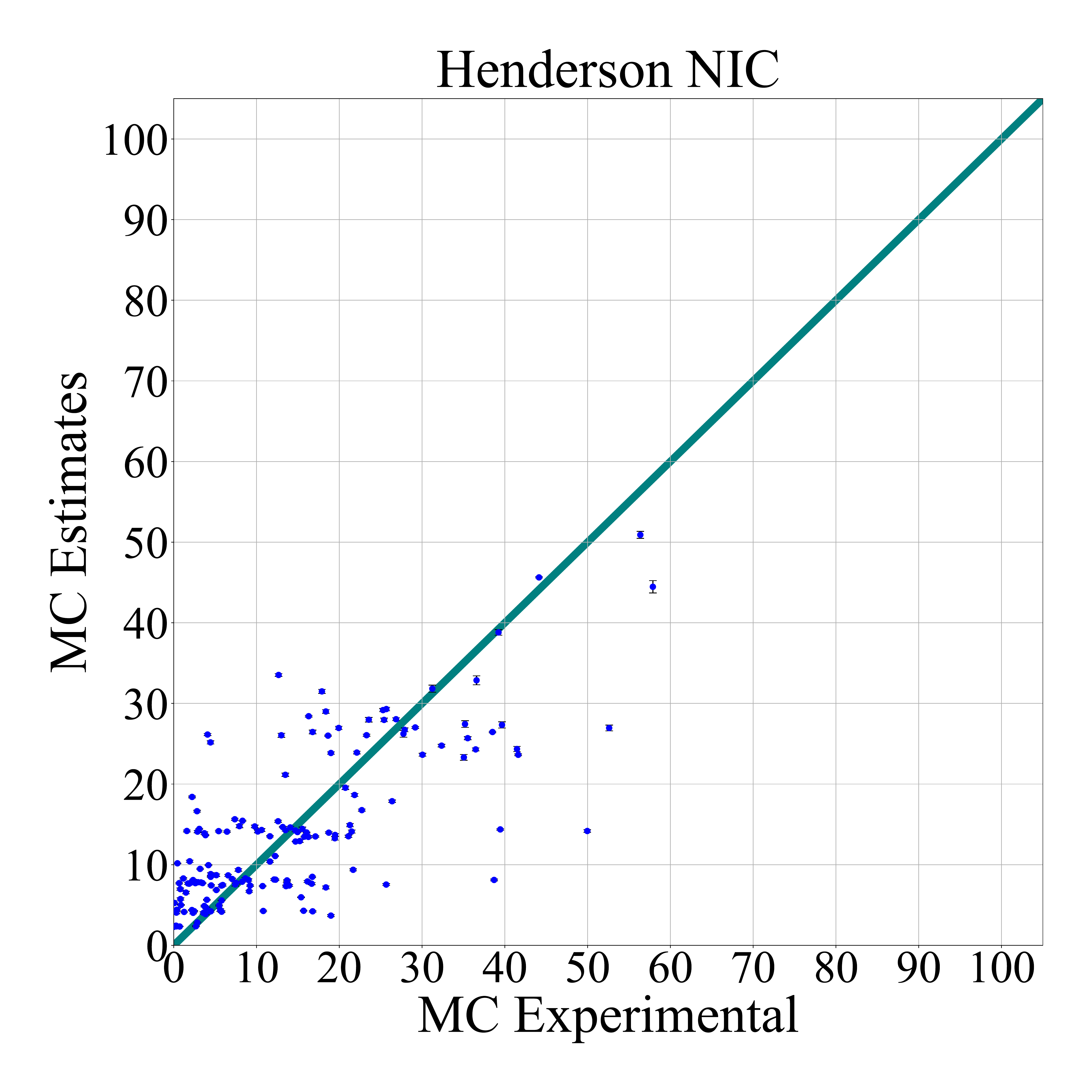} \label{fig-results-henderson_NIC_preds}} \,
    \subfloat[]{\includegraphics[width=2in]{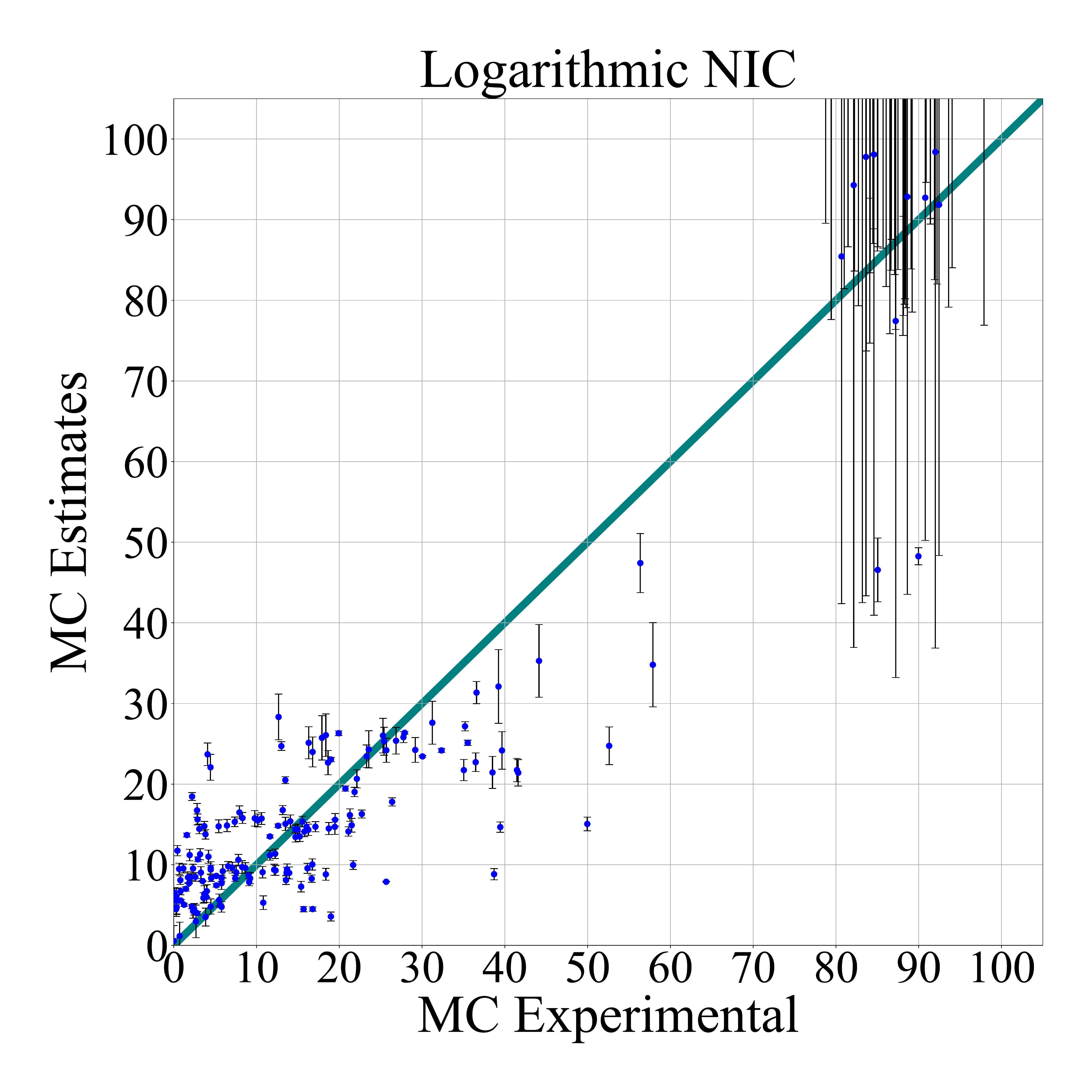} \label{fig-results-logarithmic_NIC_preds}} \,
    \subfloat[]{\includegraphics[width=2in]{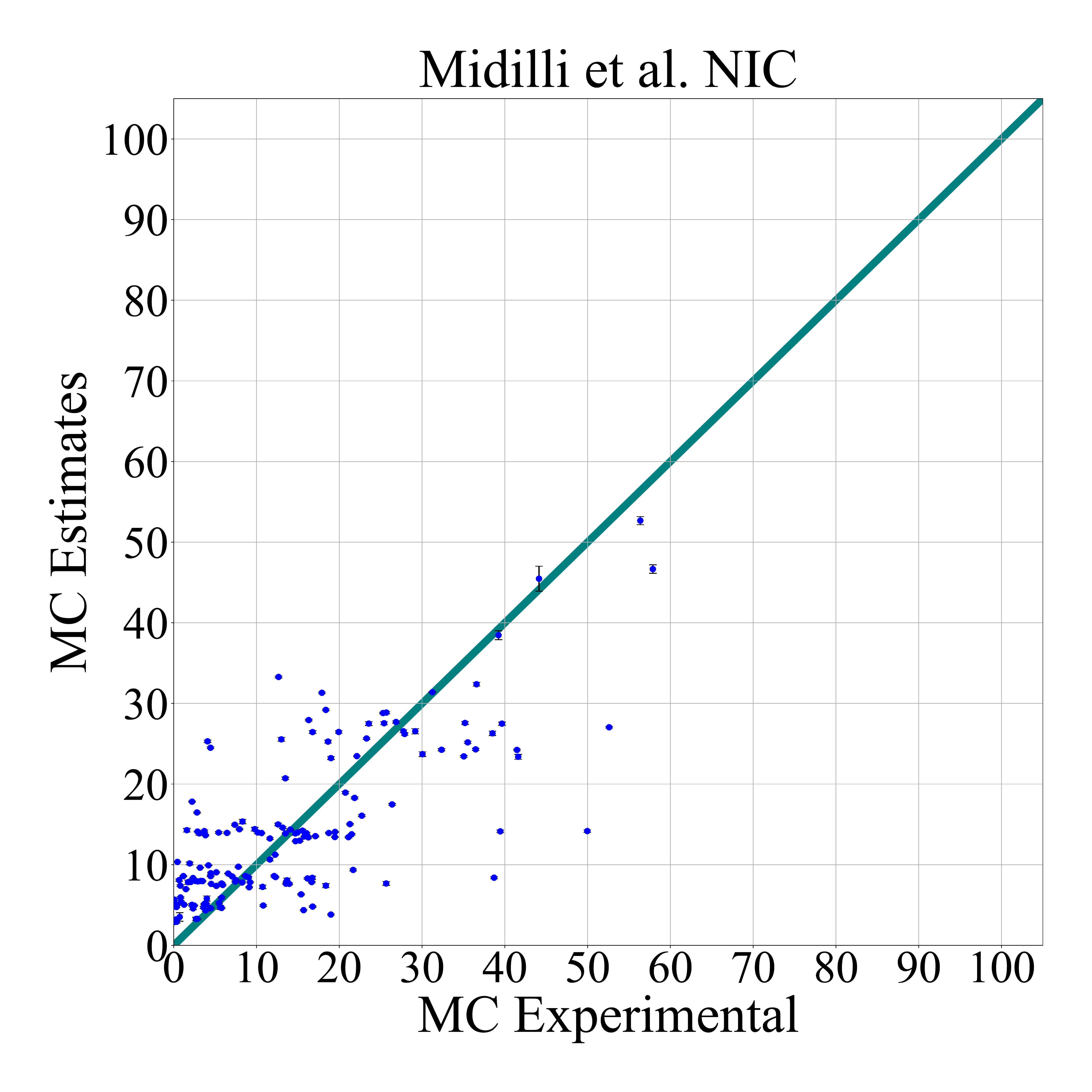} \label{fig-results-midilli_NIC_preds}} 
    \caption{Test fold results of model MC estimates for semi-empirical thin-layer drying models trained fitted using only ECD. Teal line indicates perfect estimation. Blue markers indicates the mean estimate and error bars represent the standard error of the mean, for the five times repeated cross-validation trials. Error bars are best seen on an electronic device with a zoom function. Fig. \ref{fig-results-two_term_NIC_preds}, \ref{fig-results-henderson_NIC_preds}, \ref{fig-results-logarithmic_NIC_preds}, and \ref{fig-results-midilli_NIC_preds} estimates ICD above 100 and results are thus not shown.} 
    \label{fig:Thin_layer_drying_predictions_NIC}
    \vspace{-0.5cm}
\end{figure*}

The MC estimates for the baseline thin layer drying models are shown in Fig. \ref{fig:Thin_layer_drying_predictions_WIC} and \ref{fig:Thin_layer_drying_predictions_NIC}. None of the thin layer drying models are able to satisfyingly estimate the MC along the entire drying range. The models trained on the entire data range, the WIC models, are able to correctly estimate the mean of the ICD, however the models exhibit bias as can be seen by the coefficient of determination in Table \ref{tbl:results:prediction-performance} and Fig. \ref{fig:Thin_layer_drying_predictions_WIC}. The WIC models are biased towards higher MC estimates for low MC measurements, and lower MC estimates for high experimental MC.

The Midilli et al. models as seen in Fig. \ref{fig-results-midilli_WIC_preds} and \ref{fig-results-midilli_NIC_preds}, and the Henderson NIC model Fig. \ref{fig-results-henderson_NIC_preds} are able to correctly estimate the mean of the MC of the ECD. However it is unable to deal with the large variance of the underlying data, as can be seen by the high MSE and STD measurements in Table \ref{tbl:results:prediction-performance}, thus rendering the estimation methods unsuitable for practical applications. 

The thin layer drying models fitted only on the ECD, are generally unable to estimate the ICD. The performance of MC estimation of the Lewis and Page models both improve significantly by only fitting to the ECD, whereas the more complex Two Term-, Henderson-, Logarithmic- and Midilli et al. models perform significantly better when trained including the ICD. This is as expected as these models are designed in order to improve MC estimation along the entire drying range in contrast to the older Lewis and Page models which were designed for the falling rate period.

\begin{figure*}[]
    \vspace{-0.5cm}
    \centering
    \subfloat[]{\includegraphics[width=2in]{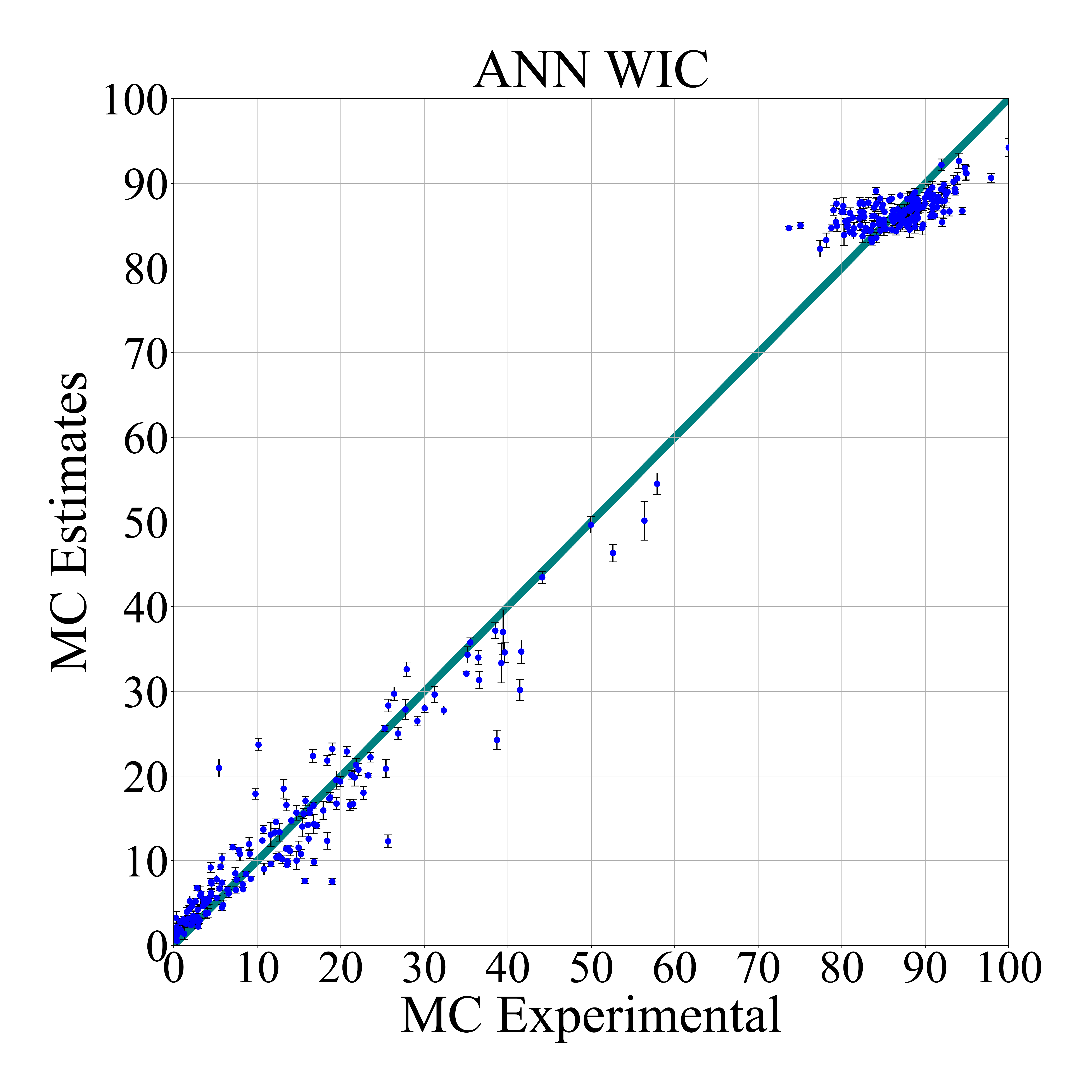} \label{fig-results-ANN_WIC_preds}}  \,
    \subfloat[]{\includegraphics[width=2in]{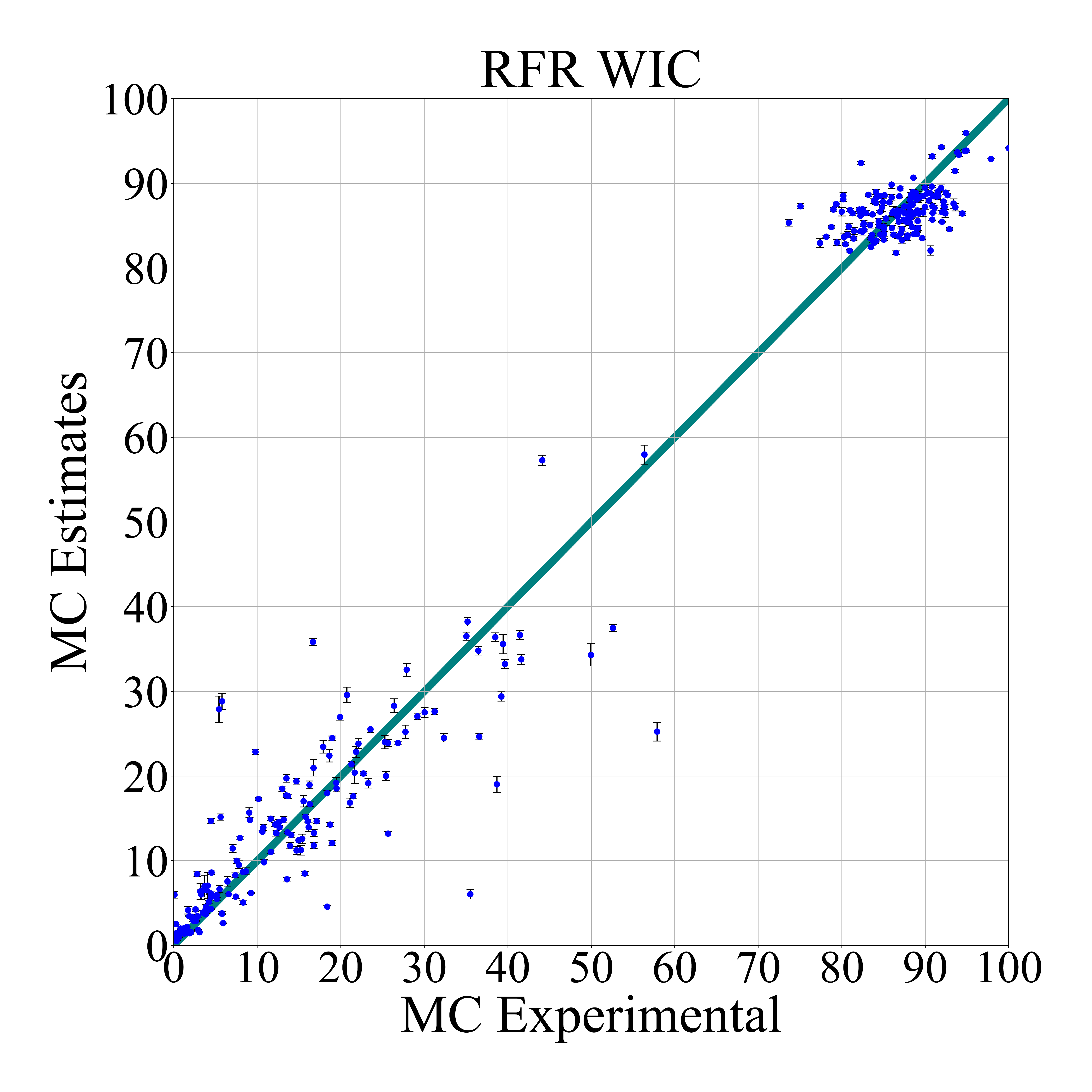} \label{fig-results-RFR_WIC_preds}} \,
    \subfloat[]{\includegraphics[width=2in]{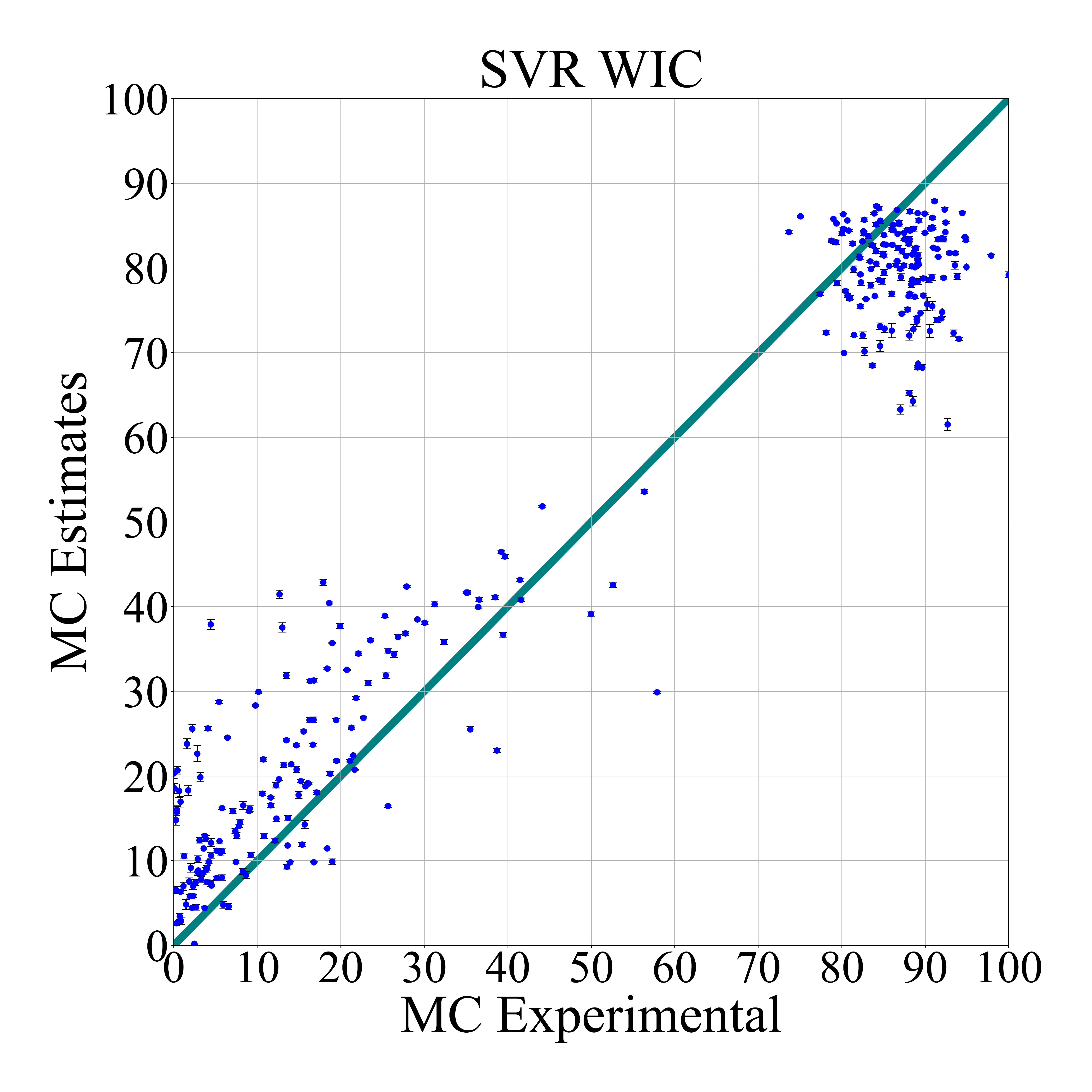} \label{fig-results-SVR_WIC_preds}}  
    \\[-2ex]
    \subfloat[]{\includegraphics[width=2in]{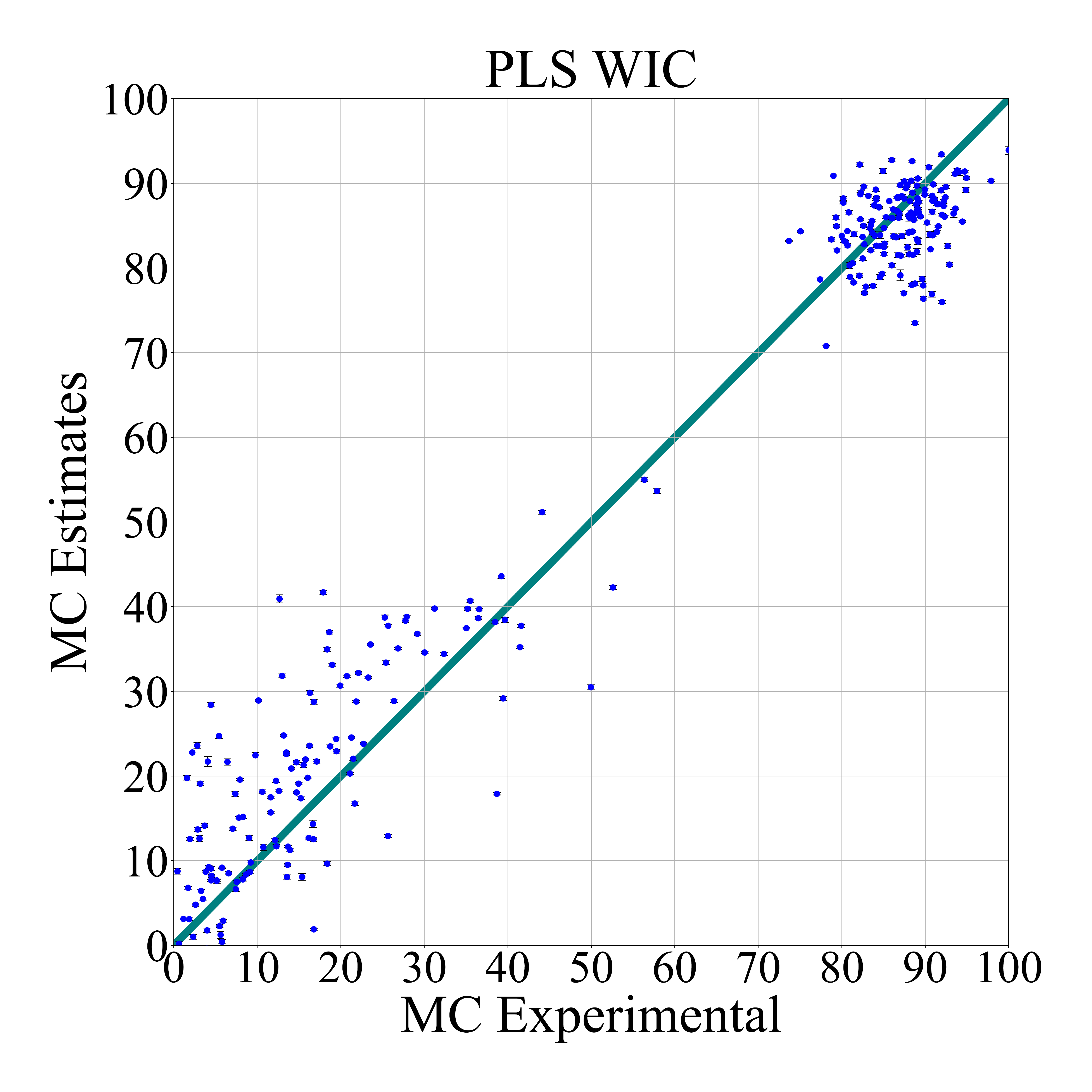} \label{fig-results-PLS_WIC_preds}} \,
    \subfloat[]{\includegraphics[width=2in]{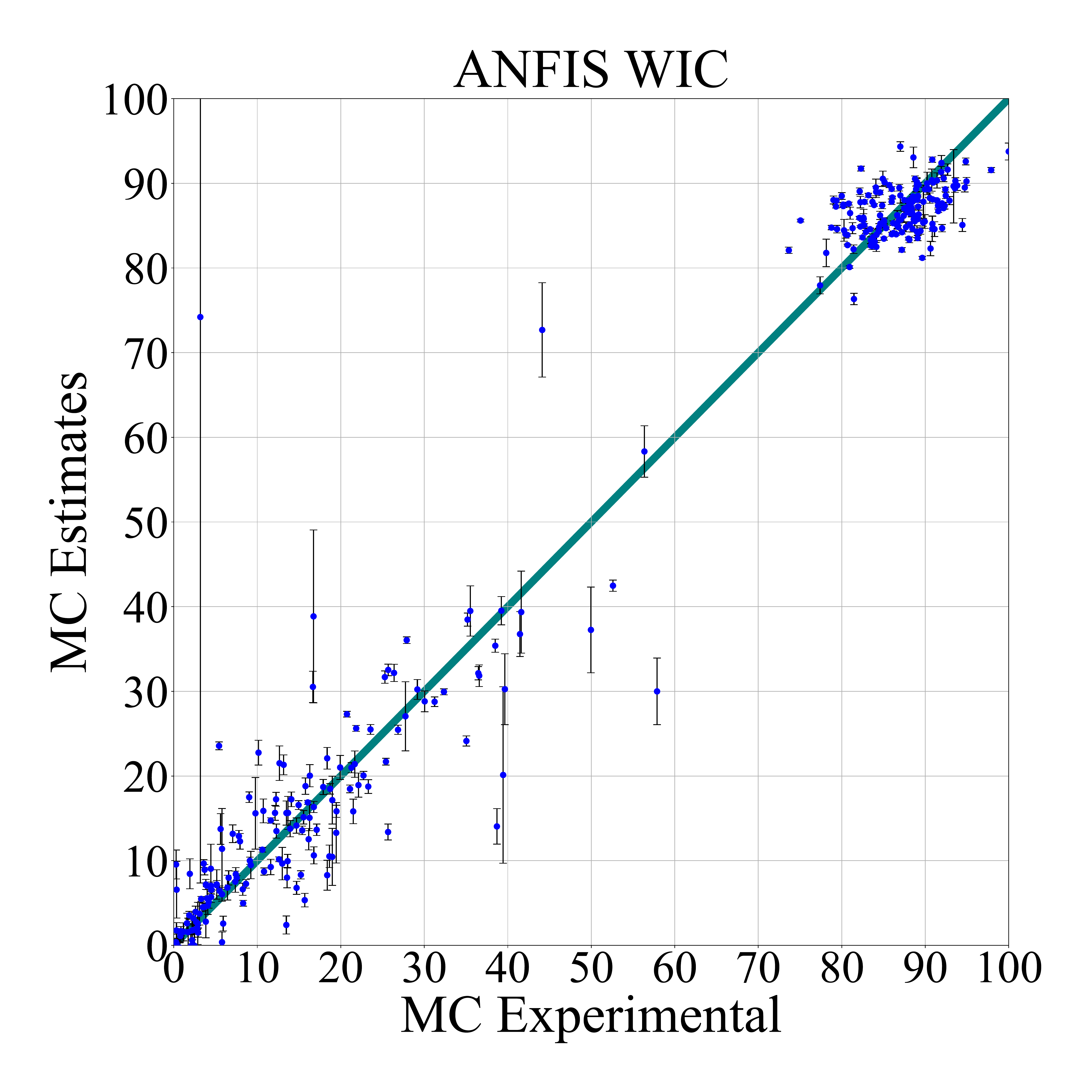} \label{fig-results-ANFIS_WIC_preds}} 
    \caption{Test fold results of model MC estimates for machine learning models trained on both ECD and ICD. Teal line indicates perfect estimation. Blue markers indicates the mean estimate and error bars represent the standard error of the mean, for the five times repeated cross-validation trials. Error bars are best seen on an electronic device with a zoom function.}
    \label{fig:data_driven-results_scatter_WIC}
    \vspace{-0.6cm}
\end{figure*}

\begin{figure*}[]
    \centering
    \subfloat[]{\includegraphics[width=2in]{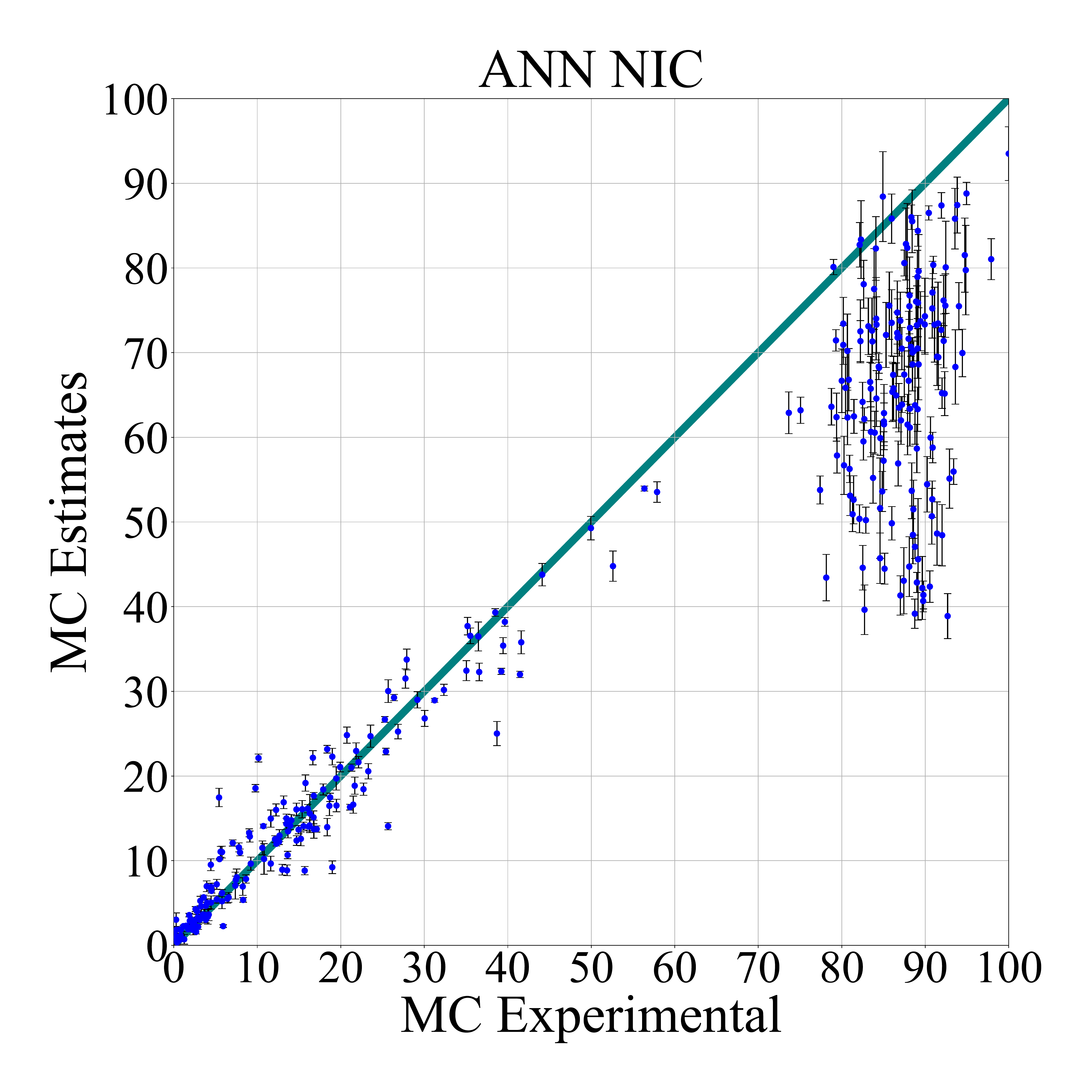} \label{fig-results-ANN_NIC_preds}}  \,
    \subfloat[]{\includegraphics[width=2in]{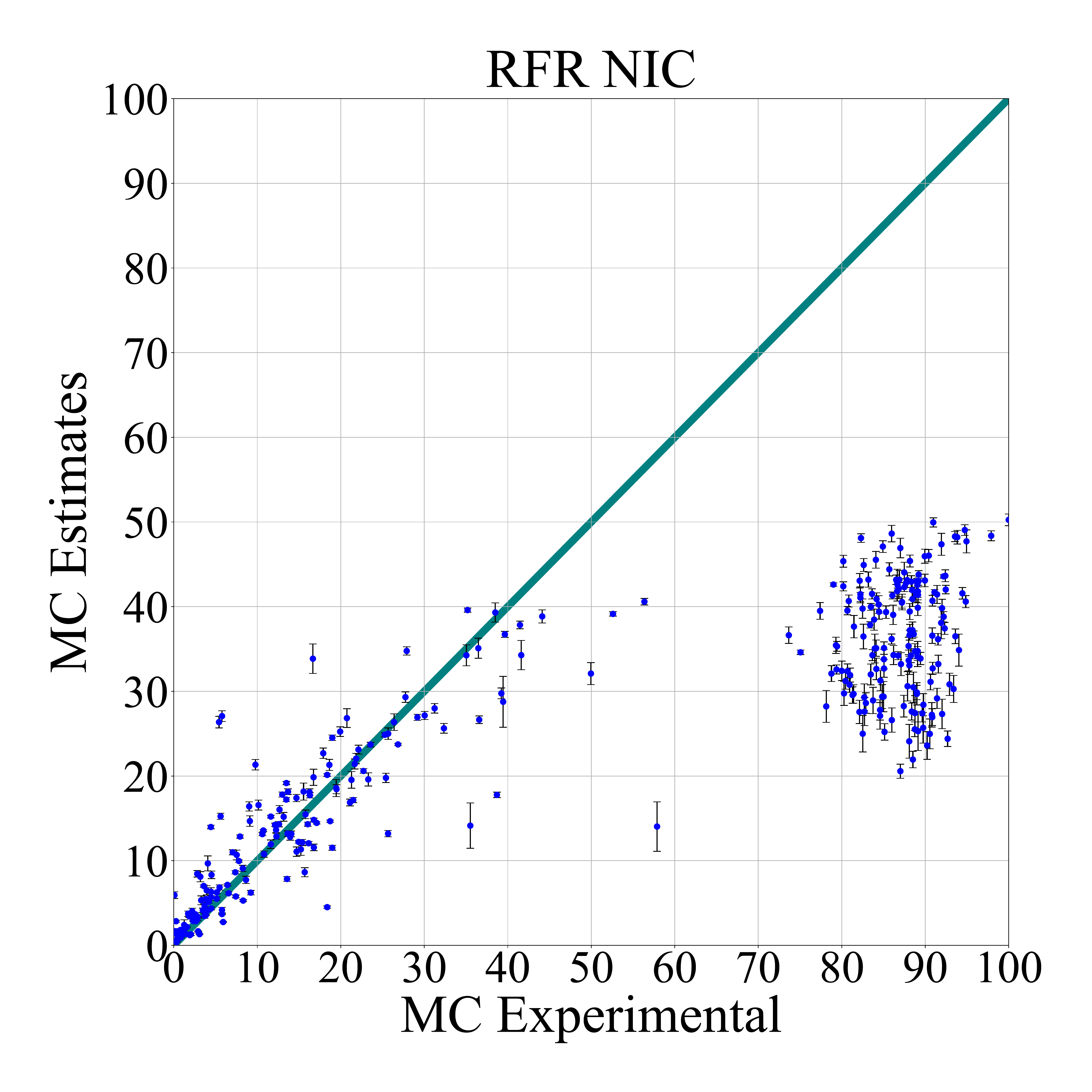} \label{fig-results-RFR_NIC_preds}} \,
    \subfloat[]{\includegraphics[width=2in]{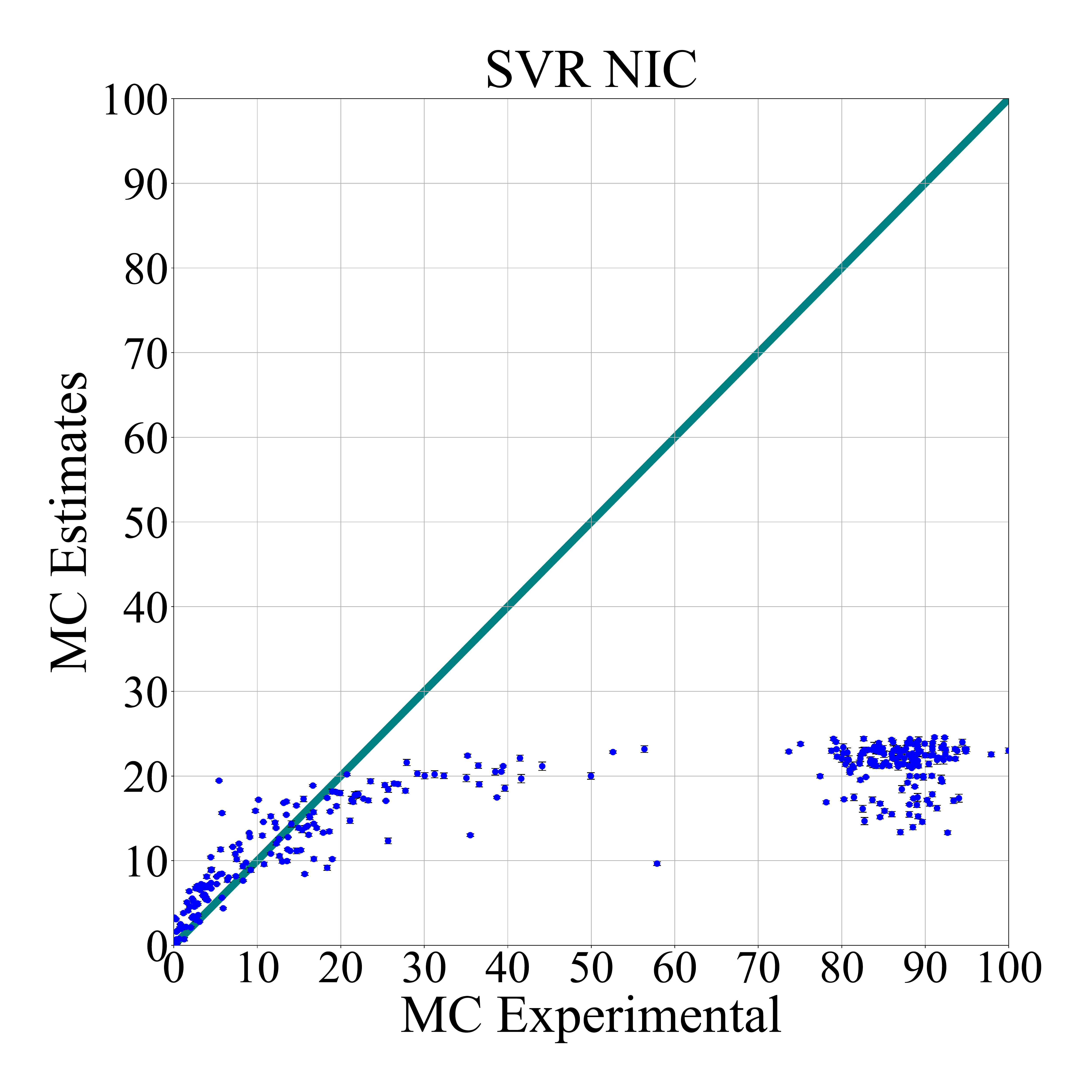} \label{fig-results-SVR_NIC_preds}}  
    \\
    \subfloat[]{\includegraphics[width=2in]{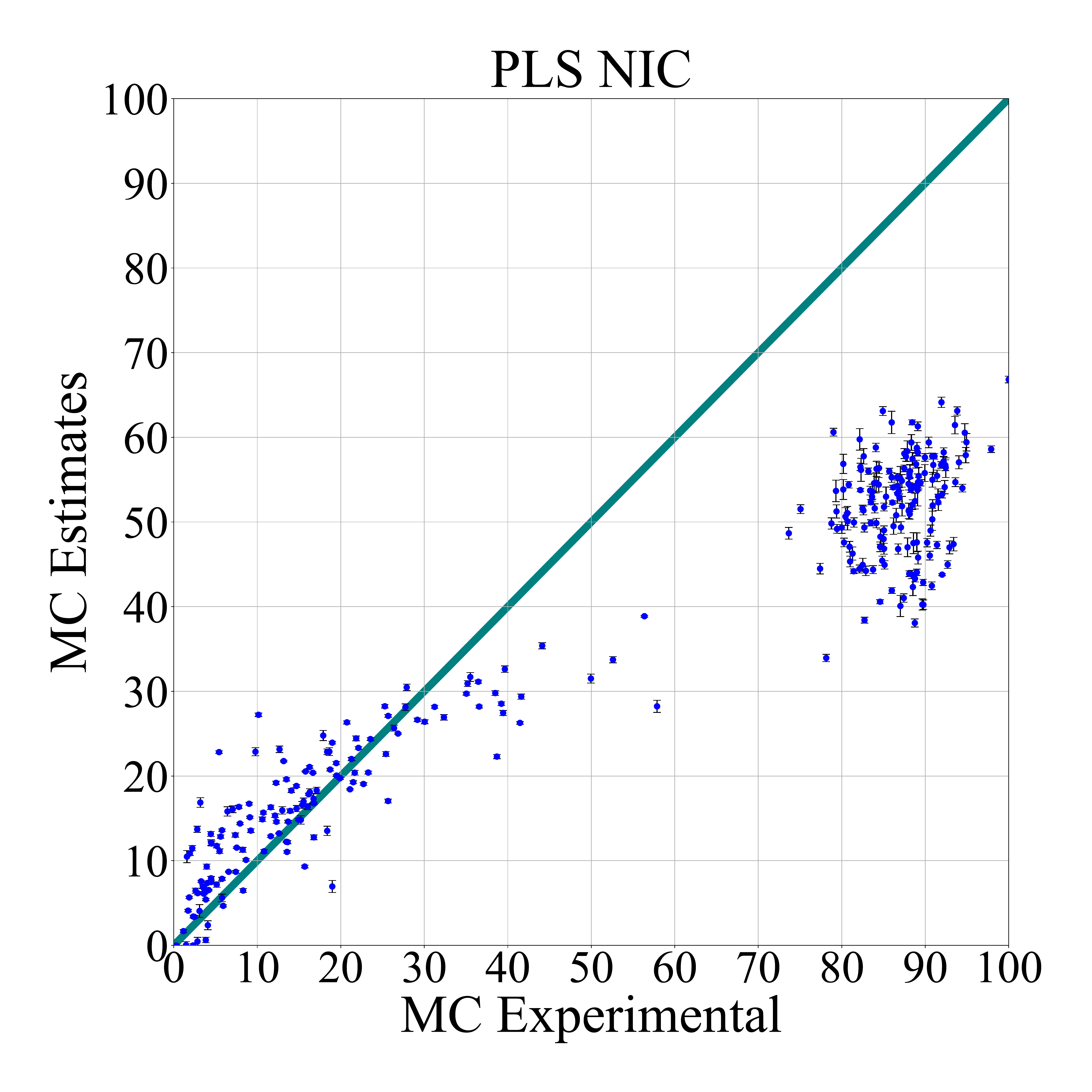} \label{fig-results-PLS_NIC_preds}} \,
    \subfloat[]{\includegraphics[width=2in]{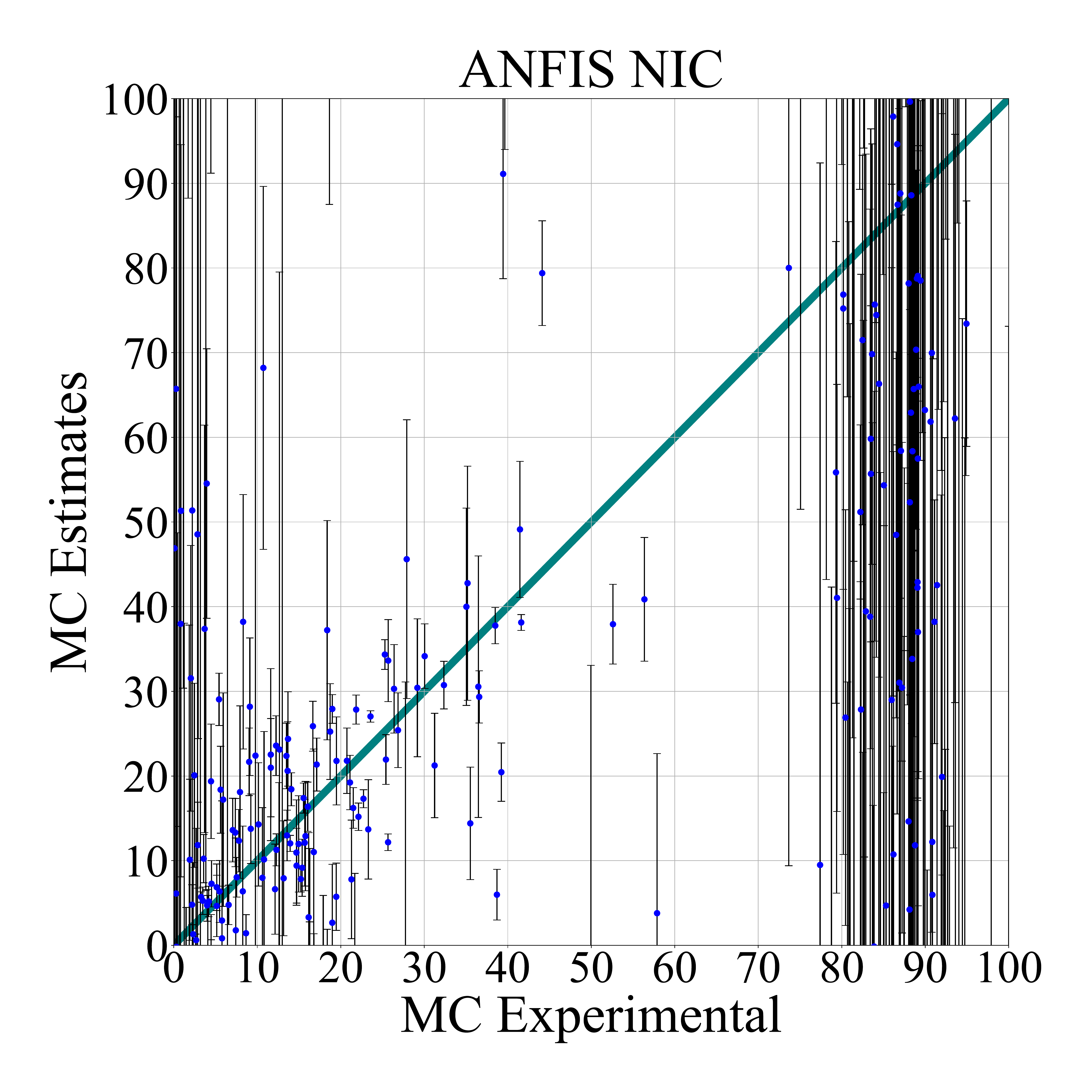} \label{fig-results-ANFIS_NIC_preds}} 
    \caption{Test fold results of model MC estimates for machine learning models trained only on ECD. Teal line indicates perfect estimation. Blue markers indicates the mean estimate and error bars represent the standard error of the mean, for the five times repeated cross-validation trials. Error bars are best seen on an electronic device with a zoom function.}
    \label{fig:data_driven-results_scatter_NIC}
    \vspace{-0.6cm}
\end{figure*}

\begin{landscape}
    \begin{table}[!h] 
    	\caption{Estimation performance of models}
    	\label{tbl:results:prediction-performance}
    	\centering
    	\begin{tabular}{p{2.5cm}llll|llll}
    		\hline \hline 
    		& \multicolumn{4}{c}{Evaluated on ECD and ICD}  & \multicolumn{4}{c}{Evaluated on ECD}\\	
    		\cmidrule(r){2-5} \cmidrule(r){6-9}
            Model 				& 	MAE $\pm$ SD			&	MSE $\pm$ SD			& STD $\pm$ SD				& $R^2$ $\pm$ SD			    & MAE $\pm$ SD     		&	MSE $\pm$ SD    		    &	STD $\pm$ SD    		& $R^2$ $\pm$ SD    	\\ \hline
                \rowcolor[HTML]{C0C0C0} Lewis WIC   &   10.0 $\pm$ 0.011		    &	137 $\pm$ 0.22	      & 6.06 $\pm$ 0.0036		    & 0.93 $\pm$ 1.3 $\cdot\text{10}^\text{-4}$  & 9.33 $\pm$ 1.2    &	82.7 $\pm$ 0.43  &	6.11 $\pm$ 0.24 & -0.90	$\pm$ 0.015	    	\\
                \rowcolor[HTML]{C0C0C0} Lewis NIC   &   10.0 $\pm$ 0.0062		    &	137 $\pm$ 0.096	      & 6.05 $\pm$ 0.0065		    & 0.93 $\pm$ $\text{5.0}\cdot\text{10}^\text{-5}$  & 6.94 $\pm$0.012  &	82.3 $\pm$ 0.19  &	5.86 $\pm$ 0.011        	    & -0.87 $\pm$	0.013	    	\\
                \rowcolor[HTML]{EFEFEF} Page WIC		&   6.57 $\pm$ 0.42	    &	81.0 $\pm$ 7.8      & 6.14 $\pm$ 0.18		    & 0.94 $\pm$ 0.0033		    & 7.16 $\pm$ 0.88  &	130 $\pm$ 0.52   &	6.32 $\pm$ 0.50     & -22.9 $\pm$ 10		    	\\
                \rowcolor[HTML]{EFEFEF} Page NIC   	&   9.64 $\pm$ 0.0051&	133 $\pm$ 0.25     & 6.34 $\pm$ 0.019		    & 0.93 $\pm$ 1.3 $\cdot\text{10}^\text{-4}$	 & 6.18 $\pm$ 0.010  &	74.5 $\pm$ 0.49  &	6.04 $\pm$ 0.039  & 0.18 $\pm$ 0.0088		    	\\
                \rowcolor[HTML]{C0C0C0} Two term WIC		    &   5.61 $\pm$ 0.14  &	58.8 $\pm$ 3.3      & 5.23 $\pm$ 0.18		    & 0.96 $\pm$ 0.0025		    & 6.03 $\pm$ 0.66  &	97.0 $\pm$ 6.6  &	5.48 $\pm$ 0.48 & 0.36 $\pm$ 0.016		    	\\
                \rowcolor[HTML]{C0C0C0} Two term NIC   	        &   -   &	 -     & - 	& -	& 6.34 $\pm$ 0.083 & 75.8 $\pm$ 1.4  & 5.98 $\pm$ 0.056   & 0.48 $\pm$ 0.051		    	\\
                \rowcolor[HTML]{EFEFEF} Henderson WIC		&  5.37 $\pm$ 0.0089   &	52.9  $\pm$ 0.15     & 4.91 $\pm$ 0.0095    & 0.96 $\pm$ 1.2 $\cdot\text{10}^\text{-4}$  & 5.66 $\pm$ 0.75   &	85.6 $\pm$ 0.28  &	5.09 $\pm$ 0.41   & -1.28 $\pm$ 0.22		    	\\
                \rowcolor[HTML]{EFEFEF} Henderson NIC   	&  82.5  $\pm$ 0.15  &	$\text{1.3}\cdot\text{10}^\text{4}$ $\pm$ 42     & 77.3 $\pm$ 0.18	    & 0.06 $\pm$ 0.0010		    & 6.12 $\pm$ 0.025   &	72.8 $\pm$ 0.25  &	5.96 $\pm$ 0.022  & 0.22 $\pm$ 0.0090	 	    	\\
                \rowcolor[HTML]{C0C0C0} Logarithmic WIC		&   5.47 $\pm$ 0.11    &	59.0 $\pm$ 1.8      & 5.39 $\pm$ 0.070	    & 0.96 $\pm$ 0.0018		    & 5.82 $\pm$ 0.78    &	97.4 $\pm$ 3.2    &	5.50 $\pm$ 0.67  & 0.23 $\pm$ 0.047		    	\\
                \rowcolor[HTML]{C0C0C0} Logarithmic NIC   	&   81.0 $\pm$ 16    &	$\text{1.6}\cdot\text{10}^\text{4}$ $\pm$ $\text{4.7}\cdot\text{10}^\text{3}$     & 95.7 $\pm$ 14	    & 0.10 $\pm$ 0.070   & 6.77 $\pm$ 0.30  &	86.9 $\pm$ 6.6  &	6.41 $\pm$0.21  & -0.3 $\pm$ 0.26		    	\\ 
                \rowcolor[HTML]{EFEFEF} Midilli et al. WIC		&   4.98 $\pm$ 0.022     &	48.1 $\pm$ 0.48      & 4.83 $\pm$ 0.030 	    & 0.96 $\pm$ 0.00033		    & 5.35 $\pm$ 0.58    &	76.0 $\pm$ 0.85    &	5.09 $\pm$ 0.54  & 0.13 $\pm$ 0.022		    	\\
                \rowcolor[HTML]{EFEFEF} Midilli et al. NIC   	&   439 $\pm$ 20    &	$\text{3.9}\cdot\text{10}^\text{5}$ $\pm$ 1.8$\cdot\text{10}^\text{4}$     & 451 $\pm$ 3.9	    & -0.65 $\pm$ 0.065   & 6.19 $\pm$ 0.032  &	73.3 $\pm$ 0.53  &	5.93 $\pm$0.024  & 0.20	$\pm$ 0.013	    	\\ \hline
                \rowcolor[HTML]{C0C0C0} ANFIS WIC		&   4.16 $\pm$ 0.18   &	46.6 $\pm$ 8.2      &    5.37 $\pm$ 0.65		    & 0.97 $\pm$ 0.0056			    & 4.92 $\pm$ 0.17      &	72.2 $\pm$ 13 	    	        &	6.90 $\pm$ 0.85        & 0.60 $\pm$ 0.067		    	\\
                \rowcolor[HTML]{C0C0C0} ANFIS NIC   	&   234 $\pm$ 94	   &	7.5$\cdot\text{10}^\text{5}$ $\pm$ 5.1 $\cdot\text{10}^\text{5}$ 	      & 766 $\pm$ 320 		 & -0.0044 $\pm$ 0.016     & 108 $\pm$ 65      & 5.2$\cdot\text{10}^\text{5}$ $\pm$ 8.5 $\cdot\text{10}^\text{5}$	       &	518 $\pm$ 493   	    & -5.3 $\cdot\text{10}^\text{-4}$ $\pm$ 0.0048  	\\
                \rowcolor[HTML]{EFEFEF} PLS WIC			&   6.15 $\pm$ 0.031	&	67.6 $\pm$ 1.0		  & 5.46 $\pm$ 0.064     & 0.95 $\pm$ 7.5$\cdot\text{10}^\text{-4}$	& 6.34 $\pm$ 0.89 &	106 $\pm$ 1.9	    	    &	5.48 $\pm$ 0.49   	& 0.61 $\pm$ 0.0044			    \\	
    		  \rowcolor[HTML]{EFEFEF} PLS NIC		  &   20.0 $\pm$ 0.058		  &	  656 $\pm$ 3.7			& 16.0 $\pm$ 0.068				  & -0.50 $\pm$ 0.010			  & 5.11 $\pm$ 0.083       		  &	  50.1 $\pm$ 2.18 	    	      &	  4.90 $\pm$ 0.16        	  & 0.59 $\pm$ 0.017		      \\
                \rowcolor[HTML]{C0C0C0} SVR WIC			&   8.25 $\pm$ 0.021			&	108 $\pm$ 1.1  & 6.32 $\pm$ 0.065	& 0.89 $\pm$ 0.0012 	& 8.23 $\pm$ 0.32  		    &	114 $\pm$ 0.83   &	6.44 $\pm$ 0.36        	& 0.25 $\pm$ 0.0068		    	\\
                \rowcolor[HTML]{C0C0C0} SVR NIC			&   35.5 $\pm$ 0.012 			&	2203 $\pm$ 2.7	  & 30.8 $\pm$ 0.033 				& -46.0 $\pm$ 0.20 		& 5.36 $\pm$ 0.023       		    &	78.5 $\pm$ 0.60 	    &	7.08 $\pm$ 0.053  	& -1.04 $\pm$ 0.014 		    	\\
                \rowcolor[HTML]{EFEFEF} RFR WIC			&   3.47 $\pm$ 0.10	&	29.3 $\pm$ 3.7		  & 4.16 $\pm$ 0.37				& 0.98 $\pm$ 2.6$\cdot\text{10}^\text{-4}$				& 3.56 $\pm$ 0.30   &	43.6 $\pm$ 7.6   &	4.48 $\pm$ 0.78        	& 0.64 $\pm$ 0.96		    	\\
                \rowcolor[HTML]{EFEFEF} RFR NIC			&   27.10 $\pm$ 0.11		&	1318 $\pm$ 9.0 	      & 24.2 $\pm$ 0.096				& -5.27 $\pm$ 0.13 				& 3.92 $\pm$ 0.11   &	45.4 $\pm$ 4.1  	    	    &	5.49 $\pm$ 0.33         	& 0.59 $\pm$ 0.030 		    	\\
                \rowcolor[HTML]{C0C0C0} ANN WIC			&  	\textbf{2.90 $\pm$ 0.050}   &	\textbf{15.1 $\pm$ 0.42}  & \textbf{2.58 $\pm$ 0.039}	& \textbf{0.99 $\pm$ 3.2$\cdot\text{10}^\text{-4}$}         & 2.83 $\pm$ 0.052 		   		    &	16.3 $\pm$ 0.50  	 		    &	\textbf{2.54 $\pm$ 0.18} 	& 0.87 $\pm$ 0.0042               \\
    		  \rowcolor[HTML]{C0C0C0} ANN NIC		  &   12.0 $\pm$ 1.1  		  &	  324 $\pm$ 51			& 13.3 $\pm$ 0.92 				  & 0.60 $\pm$ 0.091  	   		  & \textbf{2.63 $\pm$ 0.091}  	      &   \textbf{14.0 $\pm$ 1.1}       &	  2.67 $\pm$ 0.13	          & \textbf{0.90} $\pm$ 0.011	  \\	
    
    		\hline  \\
                \multicolumn{9}{l}{ - Indicates models that has not converged}
    	\end{tabular}
        \end{table}    
\end{landscape}

\subsubsection{Machine learning models}
Generally, the machine learning models as seen in Fig. \ref{fig:data_driven-results_scatter_WIC} and \ref{fig:data_driven-results_scatter_NIC} outperform the thin layer drying models. 
The ANN WIC model as seen in Fig. \ref{fig-results-ANN_WIC_preds} is able to do meaningful MC estimation on both the ICD and ECD with a low variance and low bias. Furthermore, the small standard error indicate that the model consistently performs well, independently on the selection of the cross validation folds.

The ANN NIC model as seen in Fig. \ref{fig-results-ANN_NIC_preds} does not generalize well outside the range of the data it has seen during training. However, when evaluating using only the ECD, i.e., the range of data it has been trained on, it outperforms any other model tested here on all performance metrics except for STD where it is only slightly different from the ANN WIC model. Furthermore both the ANN NIC and ANN WIC models are the only models to achieve a coefficient of determination close to one, indicating a low bias in the estimation results. 
The RFR WIC models as seen in Fig. \ref{fig-results-RFR_WIC_preds} is generally able to do the MC estimation task successfully. However it suffers from outliers, resulting in a high MSE, as seen in Table \ref{tbl:results:prediction-performance}, reducing its applicability in a real world scenario. 
Both the SVR WIC and PLS WIC models, Fig. \ref{fig-results-SVR_WIC_preds} and \ref{fig-results-PLS_WIC_preds}, generally overestimates the MC. This is preferable to underestimating the MC in a real world scenario, as it will tend to over-dry filter media instead of under-dry which is worse for the product quality. However, it is generally unable to do satisfactory MC estimation for the case of bulky filter media products. 

The performance of the ANFIS WIC and Fig. \ref{fig-results-ANFIS_WIC_preds}, can be separated into three periods. For MC $< 30$ the ANFIS WIC modes perform well, and the performance of the models are independent of the chosen folds during the 10 fold classification. However, for MC $30 < \text{MC} < 70$, it is dependent on what data was chosen for training or not, as can be seen by the standard errors indicated by the errors bars. This estimation variance originates from the tuning process of the ANFIS membership function parameters, where the entire data range has not been seen in its training folds. The ANFIS NIC model suffers badly from this problem of tuning its ANFIS membership function parameters for data outside its training folds. Fig \ref{fig-results-ANFIS_NIC_preds} shows that the ANFIS NIC model is generally unable to perform any useful estimation.
Except for the ANN NIC model, the remaining machine learning NIC models suffer from a bias towards lower estimates for higher experimental MC.

The expected estimation performance is summarized in Table \ref{tbl:results:prediction-performance} across two different ranges, one range containing both the ICD and ECD, and another range containing only the ICD. When evaluated on the both the ICD and ECD the ANN WIC model outperforms all other models on all performance criteria. When considering only the ECD, the ANN NIC model slightly outperforms the ANN WIC model on MAE, MSE and $R^2$ while the ANN WIC performs the best on the $STD$ parameter, however the difference is insignificant.

\begin{figure}[] 
	\centering
	\includegraphics[width=3in]{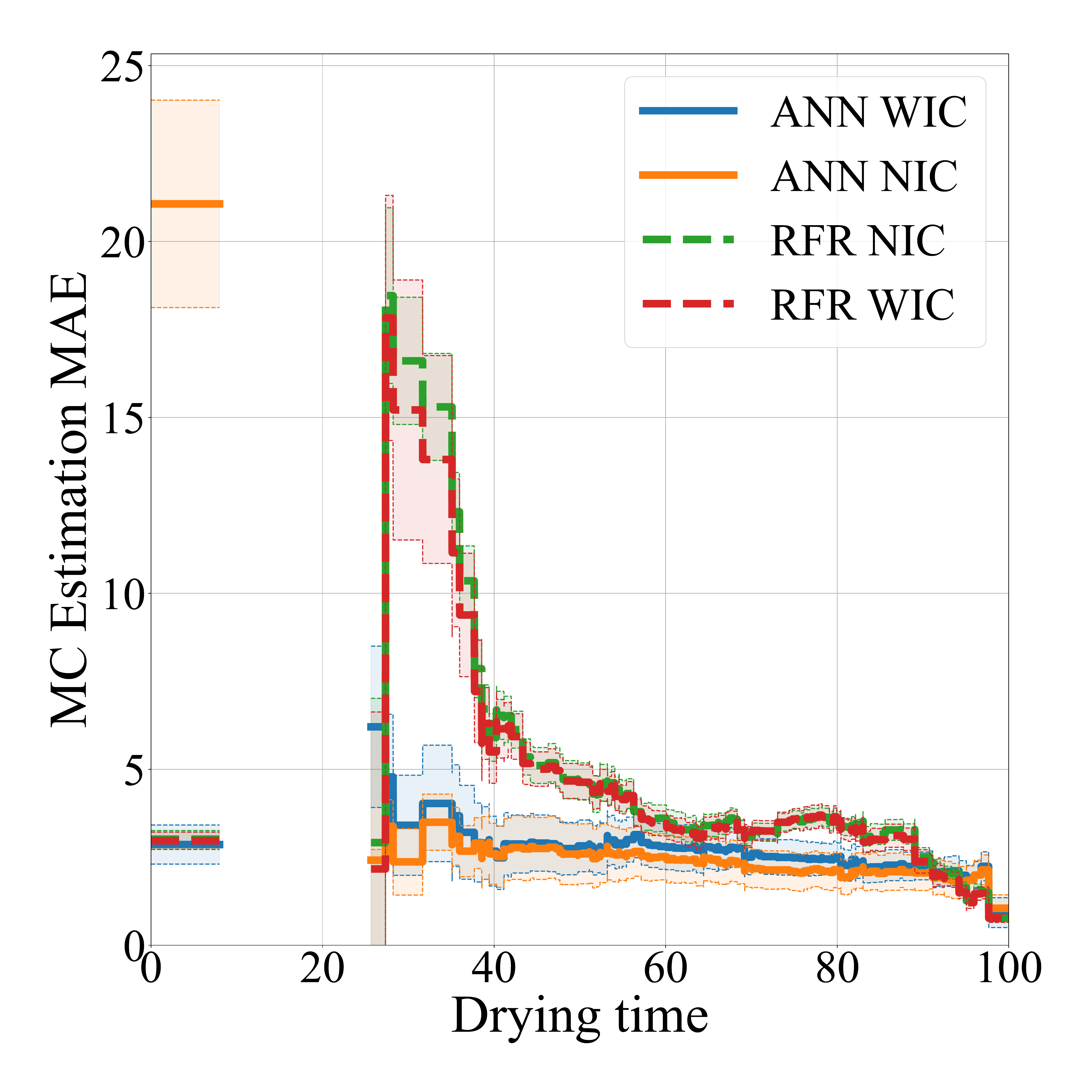} 
	\caption{Moving average of MC estimate MAE as a function of dimensionless normalized drying time with a window size of $\pm 8$. Colored areas indicate one standard deviation of the absolute residuals inside the rolling window. Area without data indicates no measurements in the area of the underlying data within the rolling window.} 
	\label{fig-results-comparison-all_models-dt-rolling_mean-vs-MAE}
\end{figure}

Taking a closer look at the four best competing models, Fig. \ref{fig-results-comparison-all_models-dt-rolling_mean-vs-MAE} shows the performance of the four best models as a function of drying time. This shows that the ANN WIC model significantly outperforms the rest of the models for the dimensionless normalized drying time of 0. The ANN WIC model seems to be able to encapsulate the drying phenomena along the entirety of the drying curve, where as RFR models are unable to correctly estimate the MC both at low and medium drying times. The ANN NIC model is unable to generalize outside of the range of the training data that it has seen. However, inside the training data range, it performs similarly or better than the ANN WIC model. All models are able to correctly estimate the MC for long drying times, however this feature is neither surprising nor interesting in a optimization setup where the goal is to decrease drying times.

We see that the estimation quality also depends on the size of the estimates. Fig. \ref{fig-results-comparison-prediction-rolling_mean-vs-MAE} shows that smaller estimates result in lower average estimation errors and standard errors. Both the average error and the standard error decreases for all models for estimates below approximately 10\% normalized MC, which is the range that is especially important for industrial applications. The high variance and mean MAE for the RFR- and ANN NIC models are caused by the models inabilities to correctly estimate the MC of the ICD. Only the ANN WIC model can correctly estimate the MC across the entire drying range.

\subsection{Practical Considerations}

\begin{figure*}[]
    \centering
    \includegraphics[width=7in]{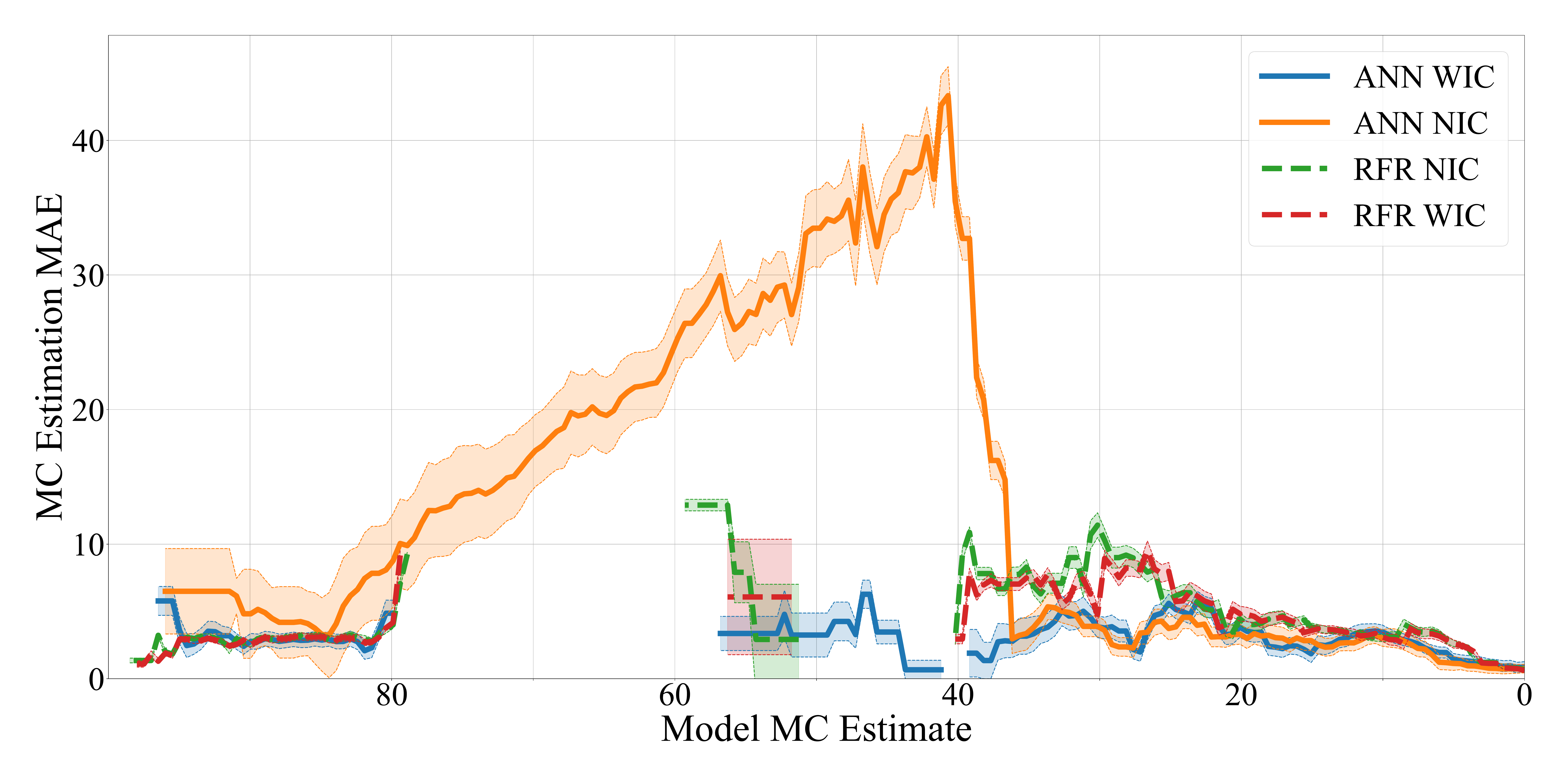}
    \caption{Model trust as defined by the moving average of MC estimate MAE as a function of model estimates with a window size of $\pm 2.5$ dimensionless normalized MC points. Colored areas indicates the standard error of the mean of the absolute residuals inside the rolling window.}
    \label{fig-results-comparison-prediction-rolling_mean-vs-MAE}
\end{figure*}

In a manufacturing setting all estimates are not made equal. The most important range of dimensionless MC estimates are below 10\%. Fig. \ref{fig-results-comparison-prediction-rolling_mean-vs-MAE} shows that in this region all models, RFR as well as ANN, perform well enough on average to be used for optimizing and controlling the manufacturing process. However the best performing model is the ANN NIC model, which can also be seen in Table \ref{tbl:results:prediction-performance}.

\section{Conclusion}\label{S:SectionV}

In this study a dataset consisting of 322 observations from 161 individual industrial drying experiments of bulky filter media products were performed and presented. 
 A total of 21 competing MC estimation models have been developed, trained, fitted, tested and compared using this data. 
The models were tested and compared using a five times repeated 10 fold cross validation scheme. 
A three layer MLP ANN was found to be the the most successful algorithm for estimating MC in bulky filter media products. The results show that including ICD to the training set of the ANN, hampers the MC estimation performance in the region of interest. This is in contrast to other investigated machine learning approaches, such as ANFIS, PLS, and SVR where either the performance increases or is unchanged when including the ICD in the training set. 
For manufacturing purposes, the most interesting region is below 10\% normalised MC as this is where one might consider stopping the drying process. Average model estimation error decreases for all models in this region, however the ANN NIC model performs the best. %
Overall, the results show that the developed ANN MC estimation approach is suitable for industrial usage. 

In general, we conclude that while thin layer drying models have been reported to be well performing for MC estimation in the fields of drying foodstuff and agricultural products, they are unable to encapsulate the underlying variance of the data of the bulky filter media. Present findings furthermore show, that ANNs combined with measurement of drying settings (oven temperature, differential pressure (fan speed) etc.) and only 2 status features, the drying time and product temperature, can be successfully used as a non-destructive MC estimation technique of bulky filter media products. Furthermore, this study establishes a baseline MC estimation performance and presents a dataset that can be used for model development and testing. As such, this study constitutes a significant contribution to both academic researchers and industrial drying designers. Further research into improving the quality of MC estimation by looking at temporal data such as change of input data is recommended. Furthermore, it could be of great interest to be able to predict the evolution of the MC, i.e. the drying curve of a specific filter media in order to be able to estimate a remaining required drying time to use for scheduling and possibly optimization of the driving oven parameters, such as oven temperature and air flow.

\section{Acknowledgement}
The authors would like to thank Cuu Van Nguyen for support, heavy lifting, and dedicated assistance during the arduous data collection phase.

\appendices

\bibliographystyle{IEEEtran}
\bibliography{IEEEabrv,bibfile}

\end{document}